\documentclass[preprint,12pt]{elsarticle}

\usepackage{comment}
\usepackage{lineno,hyperref}
\modulolinenumbers[5]
\usepackage{xfrac}
\usepackage{changepage}
\usepackage{amsfonts}
\usepackage{amsmath,amssymb}
\usepackage{graphicx, blindtext}
\usepackage{subfigure}
\usepackage{multirow}
\usepackage{multicol}
\usepackage{array}
\newcolumntype{P}[1]{>{\centering\arraybackslash}p{#1}}
\usepackage{makecell}
\usepackage{pgfplots}
\usepackage[flushleft]{threeparttable}
\usepackage{alphalph}
\usepackage{bm}
\usepackage{svg}
\definecolor{Dyellow}{RGB}{191, 144, 0}
\definecolor{Dred}{RGB}{192, 80, 77}
\definecolor{Lyellow1}{RGB}{134, 102, 0}
\definecolor{Lblue}{RGB}{43, 110, 171}
\definecolor{Lred}{RGB}{247, 59, 45}
\definecolor{Dblue1}{RGB}{60, 99, 130}
\definecolor{Dblue2}{RGB}{56, 173, 169}
\definecolor{Dred1}{RGB}{235, 47, 6}
\definecolor{Dred2}{RGB}{237, 125, 49}
\definecolor{Lblue1}{RGB}{30, 111, 166}
\definecolor{Lgray}{RGB}{105, 105, 105}
\definecolor{brown}{RGB}{114, 75, 48}
\definecolor{Lgreen}{RGB}{78, 128, 56}

\journal{ISPRS J. Photo. Remote Sens.}

\begin{document}

\begin{frontmatter}

\title{Aerial Scene Understanding in The Wild: Multi-Scene Recognition via Prototype-based Memory Networks}

 
\author[firstaddress,secondaryaddress]{Yuansheng Hua}
\ead{yuansheng.hua@dlr.de}

\author[firstaddress,secondaryaddress]{Lichao Mou}
\ead{lichao.mou@dlr.de}

\author[thirdaddress]{Jianzhe Lin}
\ead{jianzhelin@ece.ubc.ca}

\author[firstaddress,secondaryaddress]{Konrad Heidler}
\ead{konrad.heidler@dlr.de}

\author[firstaddress,secondaryaddress]{Xiao Xiang Zhu\corref{correspondingauthor}}
\cortext[correspondingauthor]{Corresponding author}
\ead{xiaoxiang.zhu@dlr.de}

\address[firstaddress]{Remote Sensing Technology Institute (IMF), German Aerospace Center (DLR),  Oberpfaffenhofen, 82234 Wessling, Germany}
\address[secondaryaddress]{Data Science in Earth Observation (SiPEO), Technical University of Munich (TUM),  Arcisstr. 21, 80333 Munich, Germany}
\address[thirdaddress]{Electrical and Computer Engineering (ECE), University of British Columbia (UBC), V6T 1Z2, Canada}

\begin{abstract}
\textit{This is a preprint. To read the final version please visit ISPRS Journal of Photogrammetry and Remote Sensing.} Aerial scene recognition is a fundamental visual task and has attracted an increasing research interest in the last few years. Most of current researches mainly deploy efforts to categorize an aerial image into one scene-level label, while in real-world scenarios, there often exist multiple scenes in a single image. Therefore, in this paper, we propose to take a step forward to a more practical and challenging task, namely multi-scene recognition in single images. Moreover, we note that manually yielding annotations for such a task is extraordinarily time- and labor-consuming. To address this, we propose a prototype-based memory network to recognize multiple scenes in a single image by leveraging massive well-annotated single-scene images. The proposed network consists of three key components: 1) a prototype learning module, 2) a prototype-inhabiting external memory, and 3) a multi-head attention-based memory retrieval module. To be more specific, we first learn the prototype representation of each aerial scene from single-scene aerial image datasets and store it in an external memory. Afterwards, a multi-head attention-based memory retrieval module is devised to retrieve scene prototypes relevant to query multi-scene images for final predictions. Notably, only a limited number of annotated multi-scene images are needed in the training phase. To facilitate the progress of aerial scene recognition, we produce a new multi-scene aerial image (MAI) dataset. Experimental results on variant dataset configurations demonstrate the effectiveness of our network. Our dataset and codes are publicly available\footnote{https://github.com/Hua-YS/Prototype-based-Memory-Network}.
\end{abstract}


\end{frontmatter}

\section{Introduction} 
\label{sec:intro}

\begin{figure}[!t]
\centering
\includegraphics[width=1\textwidth]{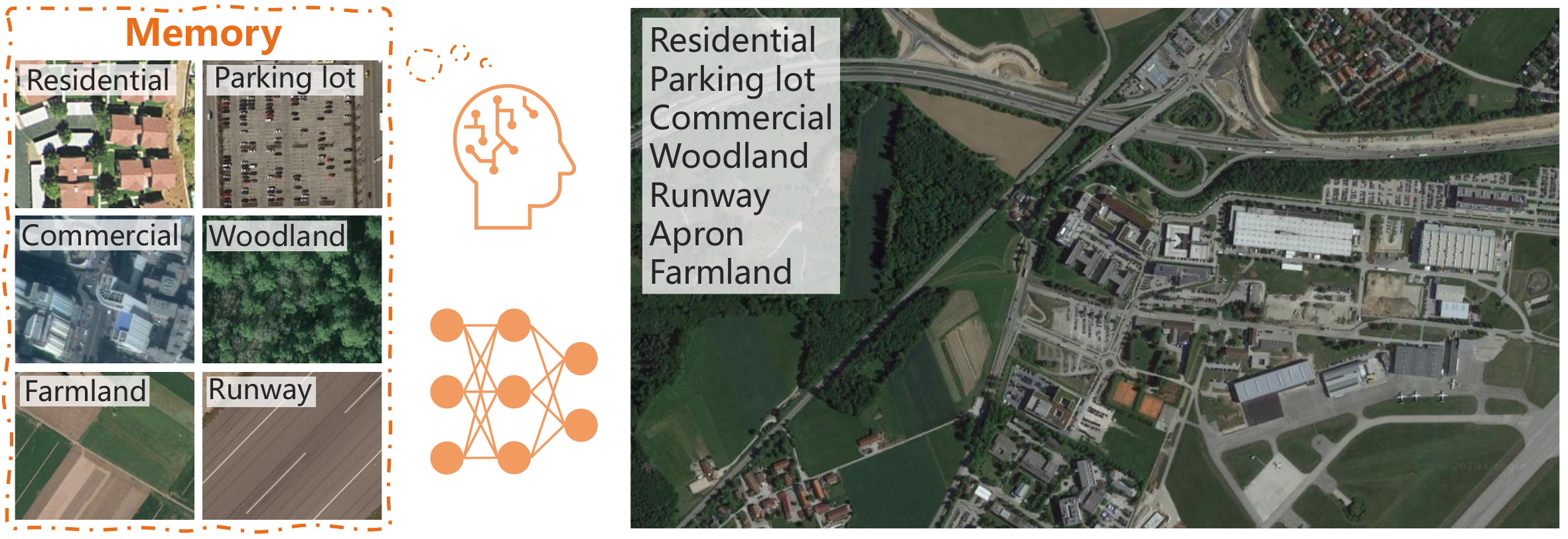}

\caption{Illustration of how humans learn to perceive unconstrained aerial images being composed of multiple scenes. We first learn and memorize individual aerial scenes. Then we can possess the capability of understanding complex scenarios by learning from only a limited number of hard instances. We believe by simulating this learning process, a deep neural network can also learn to interpret multi-scene aerial images.}

\label{fig:single_multi}
\end{figure}

With the enormous advancement of remote sensing technologies, massive high-resolution aerial images are now available and beneficial to a large variety of applications, e.g., urban planning~\cite{Marmanis17,beyongrgb,Marcos18,mou2018rifcn,hsf,qiuFCN,li2020building}, traffic monitoring~\cite{mou2018vehicle, 7729468}, disaster assessment~\cite{vetrivel2018disaster, lee2017deep}, and natural resource management~\cite{lucchesi2013applications, weng2018land, cheng2017remote, zarco2014tree, wen2017semantic, Mou18, qiu2019local}. Driven by these applications, aerial scene recognition that refers to assigning aerial images scene-level labels is now becoming a fundamental but challenging task. 

In recent years, many efforts \cite{zhu2017deep}, e.g., developing novel network architectures~\cite{murray2019zoom,cheng2020remote,bi2020multiple,niazmardi2017multiple,lin2020dual,zhu2018adaptive} and pipelines~\cite{byju2020remote,xu2020assessing,wang2019multi,zhu2019high}, publishing large-scale datasets~\cite{xia2017aid,jin2018aid++}, introducing multi-modal and multi-temporal data~\cite{hu2020cross,tuia2016multi,ru2020multi,li2020mapping}, have been deployed to address this task, and most of them treat it as a single-label classification problem. A common assumption shared by these researches is that an aerial image belongs to only one scene category, while in real-world scenarios, it is more often that there exist various scenes in a single image (cf. Figure~\ref{fig:single_multi}). Furthermore, we notice that aerial images used to learn single-label scene classification models are usually well-cropped so that target scenes could be centered and account for the majority of an aerial image. Unfortunately, this might be infeasible for practical applications. Therefore, in this paper, we aim to deal with a more practical and challenging problem, multi-scene classification in a single image, which refers to inferring multiple scene-level labels for a large-scale, unconstrained aerial image. Figure~\ref{fig:single_multi} shows an example image, where we can see that multiple scenes, e.g., \texttt{residential}, \texttt{parking lot}, and \texttt{commercial}, co-exist in one aerial image. We note that there is another research branch of aerial image understanding, multi-label object classification, which refers to the process of inferring multiple objects present in an aerial image. These studies~\cite{sumbul2019novel,zegeye2018novel,hua2020relation,khan2019graph,hua2019recurrently,zeggada2017deep,koda2018spatial} mainly focus on recognizing object-level labels, while in our task, an image is classified into multiple scene categories, which provides a more comprehensive understanding of large-scale aerial images in scene-level. To the best of our knowledge, multi-scene recognition in unconstrained aerial images still remains underexplored in the remote sensing community.

\begin{figure}[!t]
\centering
\includegraphics[width=.95\textwidth]{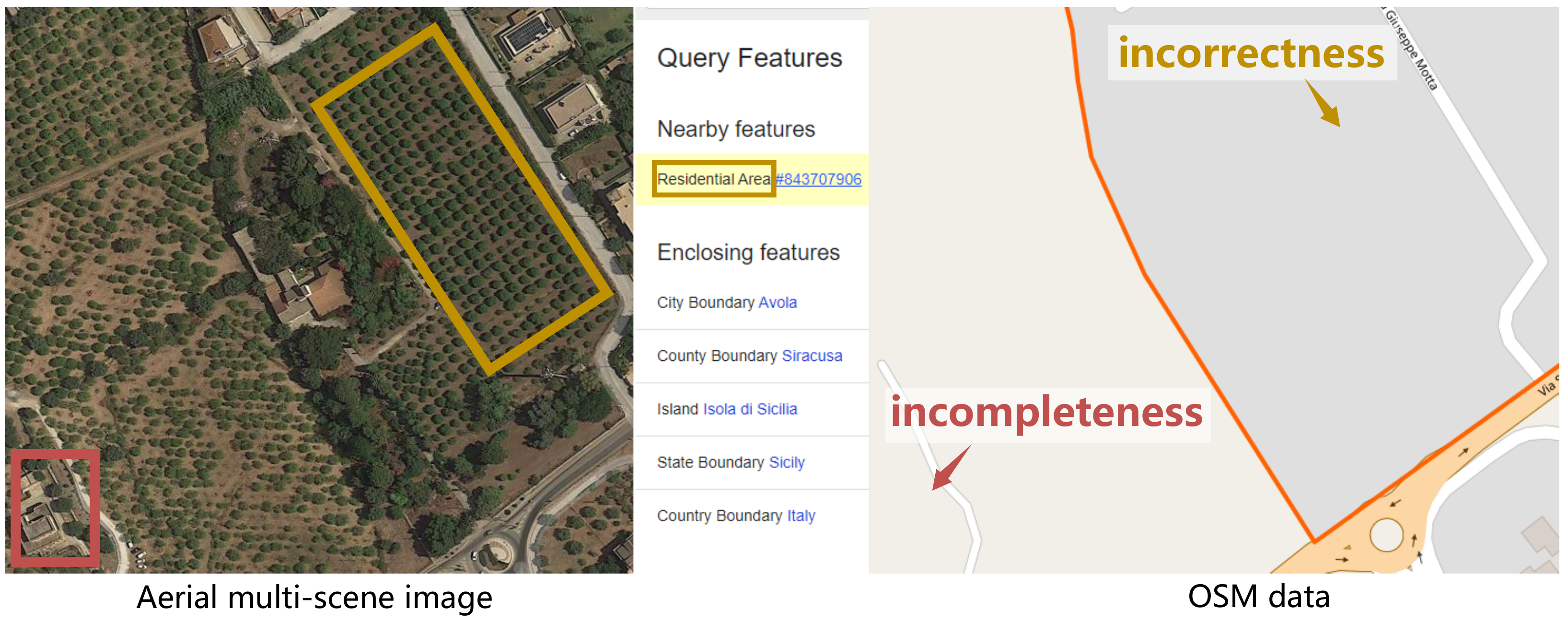}
\caption{Examples of incomplete (\textcolor{Dred}{red}) and incorrect (\textcolor{Dyellow}{yellow}) OSM data. \textcolor{Dred}{Red}: the commercial is not annotated in OSM data. \textcolor{Dyellow}{Yellow}: the orchard is mislabeled as \texttt{residential}.}
\label{fig:osm_error}
\end{figure}
To achieve this task, huge quantities of well-annotated multi-scene images are needed for the purpose of training models. However, we note that such annotations are not easy in the remote sensing community. This could be attributed to the following two reasons. On the one hand, the visual interpretation of multiple scenes is more arduous than that of a single scene in an aerial image, and therefore, labeling multi-scene images requires more work. On the other hand, low-cost annotation techniques, e.g., resorting to crowdsourcing OpenStreetMap (OSM) through keyword searching~\cite{xia2017aid, jin2018aid++,long2020dirs}, perform poorly in yielding multi-scene datasets owing to the incompleteness and incorrectness of certain OSM data. Examples of erroneous OSM data are shown in Figure~\ref{fig:osm_error}. In addition, manually rectifying annotations generated from crowdsourcing data are inevitable due to error-proneness. Such a procedure is quite labor-consuming, as every scene is required to be checked in case that present ones are mislabeled as absent. Aiming to solve the aforementioned limitations, in this work, we propose to train a network for recognizing complex multi-scene aerial images by using only a small number of labeled multi-scene images but a huge amount of existing, annotated single-scene data. Our motivation is based on an intuitive observation about how humans learn to perceive complex scenes being composed of multiple entities~\cite{national2000people,liu2008easy,mclaren2000transfer}: we first learn and memorize individual objects (through flash cards for example) when we were babies and then possess the capability of understanding complex scenarios by learning from only a limited number of hard instances (cf. Figure~\ref{fig:single_multi}). We believe that this learning process also applies to the interpretation of multi-scene aerial images. Driven by this observation, we propose a novel network, termed as prototype-based memory network (PM-Net), which is inspired by recent successes of memory networks in natural language processing (NLP) tasks~\cite{sukhbaatar2015end,miller2016key} and video analysis~\cite{shi2019dawn,park2020learning,lai2020mast}. To be more specific, we first learn the prototype representation of each aerial scene from single-scene aerial images and then store these prototypes in the external memory of PM-Net. Afterwards, for a given query multi-scene image, a multi-head attention-based memory retrieval module is devised to retrieve scene prototypes that are associated with the query image from the external memory for inferring multiple scene labels.
\par
The contributions of this work are fourfold.
\begin{itemize}
    \item We take a step forward to a more practical and challenging task in aerial scene understanding, namely multi-scene classification in single images, which aims to recognize multiple scenes present in a large-scale, unconstrained aerial image. Such a task is in line with real-world scenarios and capable of providing a comprehensive picture for a given geographic area.
    \item Given that labeling multi-scene images is very labor-intensive and time-consuming, we propose a PM-Net that can be trained for our task by leveraging large numbers of existing single-scene aerial images and a small number of labeled multi-scene images.
    \item In order to facilitate the progress of multi-scene recognition in single aerial images, we create a new dataset, multi-scene aerial image (MAI) dataset. To the best of our knowledge, this is the first publicly available dataset for aerial multi-scene interpretation. Compared to existing single-scene aerial image datasets, images in our dataset are unconstrained and contain multiple scenes, which are more in line with the reality.
    \item We carry out extensive experiments with different configurations. Experimental results demonstrate the effectiveness of the proposed network.
\end{itemize}

The remaining sections of this paper are organized as follows. Section~\ref{sec:relatedwork} reviews studies in memory networks and prototypical networks, and the architecture of the proposed prototype-based memory network is introduced in Section~\ref{sec:method}. Section~\ref{sec:experiment} describes experimental configurations and analyzes results. Eventually, conclusions are drawn in Section~\ref{sec:conclusion}.

\section{Related Work}
\label{sec:relatedwork}
Since very few efforts have been deployed to this task in the remote sensing community, we only review literatures related to our algorithm in this section.

\subsection{Memory Networks}
A memory network takes as input a query and retrieves complementary information from the external memory. In~\cite{sukhbaatar2015end}, the memory network is first proposed and utilized to address question-answering tasks, where questions are regarded as queries, and statements are stored in the external memory. To retrieve statements for predicting answers, the authors compute relative distances between queries and the external memory through dot product. In the following work, Miller et al. ~\cite{miller2016key} improves the efficiency of retrieving large memories by pre-selecting small subsets with key hashing. Moreover, the memory network is further applied in video analysis~\cite{shi2019dawn,park2020learning,lai2020mast} and image captioning~\cite{cornia2020meshed}. In~\cite{shi2019dawn}, the authors devise a dual augmented memory network to memorize both target and background features of an video, and use a Long Short-Term Memory (LSTM) to communicate with previous and next frames. In~\cite{park2020learning}, the authors propose a memory network to memorize normal patterns for detecting anomalies in an video. As an attempt in image captioning, Cornia et al.~\cite{cornia2020meshed} devise a learnable memory to learn and memorize priori knowledge for encoding relationships between image regions. Inspired by these works, we devise a memory network and store scene prototypes in the memory for recognizing scenes present in multi-scene images.

\subsection{Prototypical Networks}
Prototypical networks are characterized by classifying images according to their distances from class prototypes. In learning with limited training samples, such networks are popular and achieved many successes recently~\cite{snell2017prototypical,guerriero2018deepncm,yang2018robust,huang2020relational,zhang2020global,tang2019spatial}. To be specific, Snell et al.~\cite{snell2017prototypical} propose to first learn a prototype representation for each category and then identify images by finding their nearest category prototypes. Guerriero et al.~\cite{guerriero2018deepncm} aim to alleviate the heavy expense of learning prototypes by initializing and updating prototypes with those learned in previous training epochs. Yang et al.~\cite{yang2018robust} propose to combine prototypical networks and CNNs for tackling the open world recognition problem and improving the robustness and accuracy of networks. Similarly, Huang et al.~\cite{huang2020relational} propose to integrate prototypical networks and graph convolutional neural networks for learning relational prototypes. Albeit variant, most existing works share a common way to extract prototypes, which is taking average of samples belonging to the same categories. Therefore, we follow this prototype extraction strategy in our work.

\section{Methodology}
\label{sec:method}

\begin{figure*}[!t]
\centering
\includegraphics[width=1\textwidth]{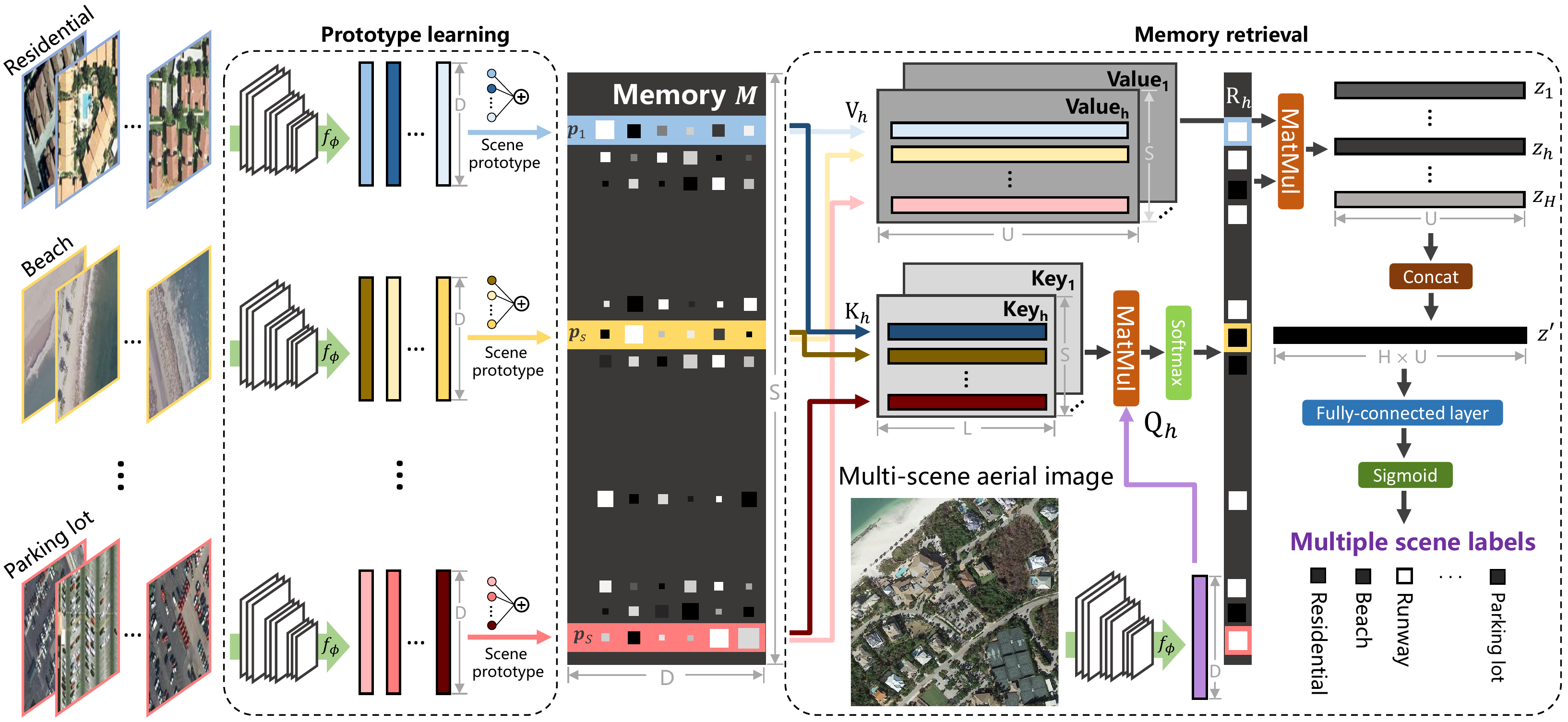}
\caption{Architecture of the proposed PM-Net. Particularly, we first learn scene prototypes $\bm{p}_s$ from well-annotated single-scene aerial images and then store them in the external memory $\bm{M}$ of PM-Net. Afterwards, given a query multi-scene image, a multi-head attention-based memory retrieval module is devised to retrieve scene prototypes that are relevant to the query image, yielding $\bm{z}'$ for the prediction of multiple labels. $f_\phi$ denotes the embedding function, and its output is a $D$-dimensional feature vector. $S$ and $H$ represent numbers of scenes and heads, respectively. $L$ and $U$ denote channel dimensions of the key and value in the memory retrieval module.}
\label{fig:samp2}
\end{figure*}

\subsection{Overview}
The proposed PM-Net consists of three essential components: a prototype learning module, an external memory, and a memory retrieval module. Specifically, the prototype learning module is devised to encode prototype representations of aerial scenes, which are then stored in the external memory. The memory retrieval module is responsible for retrieving scene prototypes related to query images through a multi-head attention mechanism. Eventually, retrieved scene prototypes are utilized to infer the existence of multiple scenes in the query image.

\subsection{Scene Prototype Learning and Writing}
\label{sec:learn_sp}

Following the observation introduced in Section \ref{sec:intro}, we propose to learn and memorize scene prototypes with the support of single-scene aerial images. The procedure consists of two stages. We first employ an embedding function to learn semantic representations of all single-scene images. Then, feature representations belonging to the same scene category are encoded into a scene prototype and stored in the external memory.

Formally, let $\bm{X}^s_{i}$ denote the $i$-th single-scene image belonging to scene $s$, and $i$ ranges from $1$ to $N_s$. $N_s$ is the number of samples annotated as $s$. The embedding function $f_{\phi}$ can be learned via the following objective function:
\begin{equation}
\label{eq:cross_entropy}
    \mathcal{L}(\bm{X}^s_{i}, \bm{y}^s) = -\bm{y}^s \log{ \frac{\exp{(-g_\theta(f_\phi(\bm{X}^s_{i})))}}{\sum\limits_s \sum\limits_{i} \exp{(-g_\theta(f_\phi(\bm{X}^s_{i})))}}},
\end{equation}
where $\phi$ represents learnable parameters of $f_{\phi}$, and $\bm{y}^s$ is a one-hot vector denoting the scene label of $\bm{X}^s_{i}$. $g_\theta$ is a multilayer perceptron (MLP) with parameters $\theta$ and its outputs are activated by a softmax function to predict probability distributions. Following the overwhelming trend of deep learning, here we employ a deep CNN, e.g., ResNet-50~\cite{he2016deep}, as the embedding function $f_\phi$ and learn its parameters on public single-scene aerial image datasets. After sufficient training, $f_\phi$ is expected to be capable of learning discriminative representations for different aerial scenes. 

Once $f_{\phi}$ is learned, the scene prototype can be computed by averaging representations of all aerial images belonging to the same scene~\cite{snell2017prototypical,guerriero2018deepncm,yang2018robust}. Let $\bm{p}_s$ be the prototype representation of scene $s$. We calculate $\bm{p}_s$ with the following equation:
\begin{equation}
\label{eq:mean_ps}
    \bm{p}_s = \frac{1}{N_s}\sum_{i=1}^{N_s} f_\phi(\bm{X}^s_{i}).
\end{equation}

By doing so, in the single-scene classification, an image closely around $\bm{p}_s$ in the common embedding space is supposed to belong to scene $s$. Similarly, in the multi-scene scenario, the representation of an aerial image comprising scene $s$ should show high relevance with $\bm{p}_s$. After encoding all scene prototypes, the external memory $\bm{M}$ can be formulated as follows:
\begin{equation}
\label{eq:mem}
    \bm{M}=[\bm{p}_{1}, \bm{p}_{2}, ..., \bm{p}_{S}]^T,
\end{equation}
where $S$ denotes the number of scenes. $[\cdots, \cdots]$ represents the concatenation operation. Given that $\bm{p}_s$ is a $D$-dimensional vector, $\bm{M}$ is a matrix of $S \times D$. Note that $D$ varies when using different backbone CNNs as embedding functions.

\subsection{Multi-head Attention-based Memory Retrieval}

Inspired by successes of the multi-head self-attention mechanism~\cite{vaswani2017attention} in natural language processing tasks~\cite{radford2018improving,Radford2019LanguageMA,devlin2018bert,wolf2020transformers}, we develop a multi-head attention-based memory retrieval module to retrieve scene prototypes from the memory $\bm{M}$ for a given query image $\bm{X}$. Given a query multi-scene aerial image $\bm{X}$, to retrieve relevant scene prototypes from $\bm{M}$, we develop a multi-head attention-based memory retrieval module. In particular, we first extract the feature representation of $\bm{X}$ through the same embedding function $f_\phi$ and linearly project it to an $L$-dimensional query ${\rm Q}(\bm{X})$. Similarly, we transform the external memory $\bm{M}$ into key ${\rm K}(\bm{M})$ and value ${\rm V}(\bm{M})$, and both are implemented as MLPs. The channel dimension of the key is $L$, while that of the value is $U$. The relevance between $\bm{X}$ and each scene prototype $\bm{p}_s$ can be measured by dot product similarity and a softmax function as follows:
\begin{equation}
\label{eq:attention}
    {\rm R}(\bm{X}, \bm{M})={\rm softmax}(\frac{{\rm Q}(f_\phi(\bm{X}))\cdot {\rm K}(\bm{M})^T}{\sqrt{L}}).
\end{equation}

The output is an $S$-dimensional vector, where each component represents a relevance probability that a specific scene prototype is related to the query image. Subsequently, the retrieved scene prototypes are computed by weight-summing all values with the following equation:
\begin{equation}
\label{eq:v_a}
    \bm{z}={\rm R}(\bm{X}, \bm{M})\cdot {\rm V}(\bm{M}).
\end{equation}

Since the memory retrieval is designed in a multi-head fashion, the final retrieved prototype is reformulated as follows:
\begin{equation}
\label{eq:m_r}
\bm{z}' = [\bm{z}_1, \bm{z}_2, ..., \bm{z}_H],
\end{equation}
where $H$ denotes the number of heads, and each head yields a retrieved prototype $\bm{z}_h$ by transforming $\bm{X}$ and $\bm{M}$ to the variant query ${\rm Q}_h(f_\phi(\bm{X}))$, key ${\rm K}_h(\bm{M})$, and value ${\rm V}_h(\bm{M})$. Eventually, the output $\bm{z}'$ is fed into a fully-connected layer followed by a sigmoid function for inferring presences of aerial scenes.

\subsection{Implementation Details}
For a comprehensive assessment of our PM-Net, we implement the embedding function with various backbone CNNs. Specifically, we conduct experiments on four CNN architectures, and details are as follows:
\begin{itemize}
    \item PM-VGGNet: $f_\phi$ is built on VGG-16~\cite{simonyan2014very} by replacing all layers after the last max-pooling layer in \textit{block5} with a global average pooling layer.
    \item PM-Inception-V3: Inception-V3~\cite{szegedy2015going} is utilized, and layers before and including the global average pooling layer are employed as $f_\phi$.
    \item PM-ResNet: We modify ResNet-50~\cite{he2016deep} by discarding layers after the global average pooling layer and using the remaining layers as $f_\phi$.
    \item PM-NASNet: The backbone of $f_\phi$ is mobile NASNet~\cite{nas}. As with the modification in PM-ResNet, only layers before and including the global average pooling layer are used.
\end{itemize}

In our experiments, we train original deep CNNs on single-scene aerial image datasets and then take them as the embedding function $f_\phi$ following the aforementioned points. Subsequently, we yield scene prototypes $\bm{p}_s$ and concatenate all of them along the first axis to form $\bm{M}$.


\section{Experiments and Discussion}
\label{sec:experiment}
In this section, we introduce a newly produced multi-scene aerial image dataset, MAI dataset, and two single-scene datasets, i.e., UCM and AID datasets, which are used in experiments. Then network configurations and training schemes are detailed in Subsection \ref{sec:train}. The remaining subsections discuss and analyze the performance of the proposed network thoroughly.

\begin{figure*}[!t]
\centering
\subfigure[]{\includegraphics[width=0.3\textwidth]{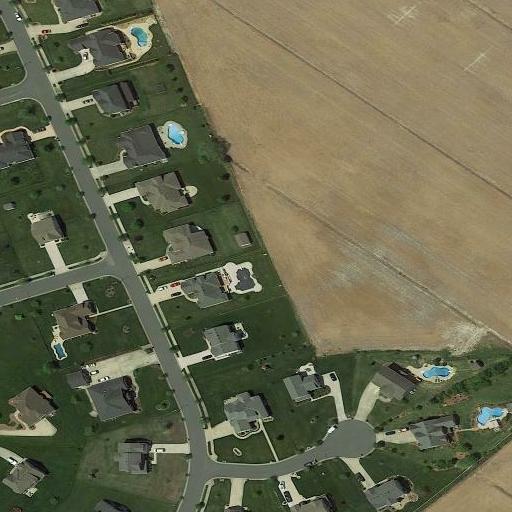}
\label{fig:le1}}
\subfigure[]{\includegraphics[width=0.3\textwidth]{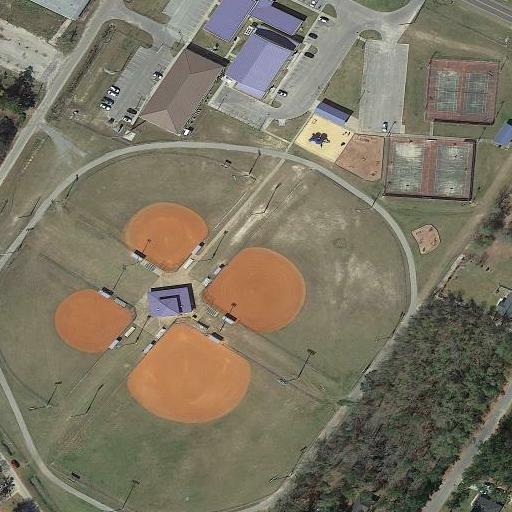}
\label{fig:le2}}
\subfigure[]{\includegraphics[width=0.3\textwidth]{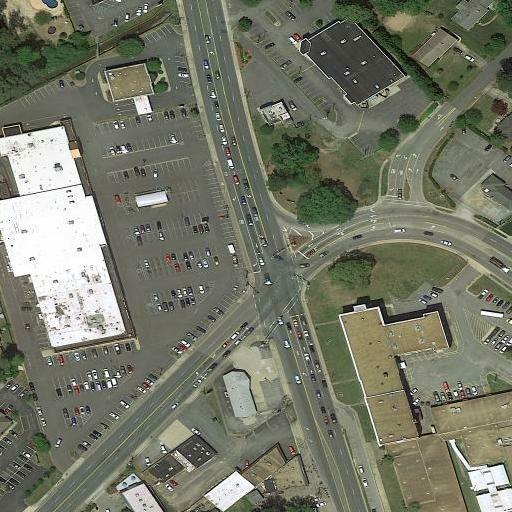}
\label{fig:le3}}
\vspace{-0.15cm}
\subfigure[]{\includegraphics[width=0.3\textwidth]{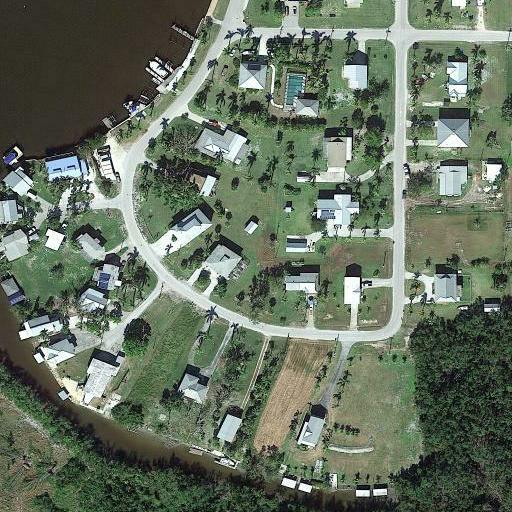}
\label{fig:le4}}
\subfigure[]{\includegraphics[width=0.3\textwidth]{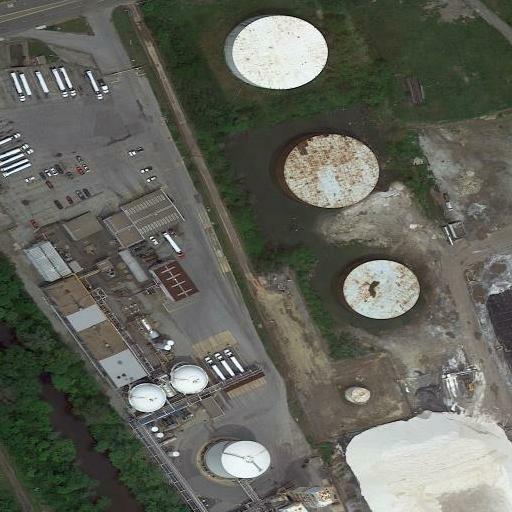}
\label{fig:le15}}
\subfigure[]{\includegraphics[width=0.3\textwidth]{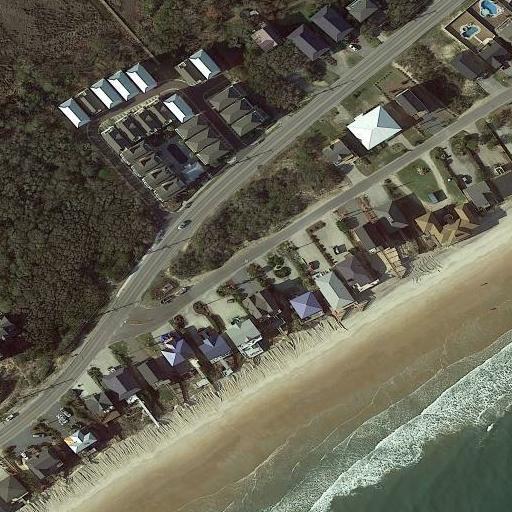}
\label{fig:le6}}
\vspace{-0.15cm}
\subfigure[]{\includegraphics[width=0.3\textwidth]{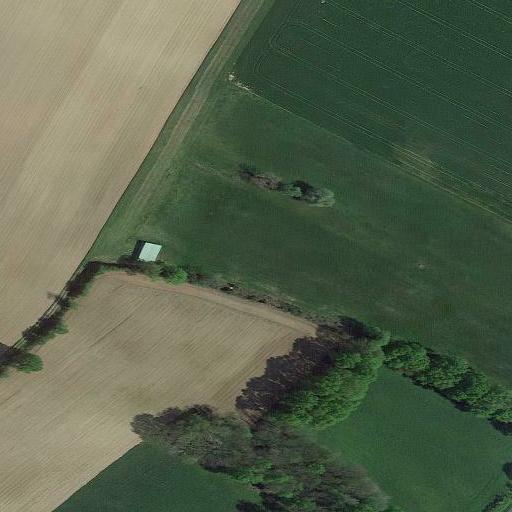}
\label{fig:le7}}
\subfigure[]{\includegraphics[width=0.3\textwidth]{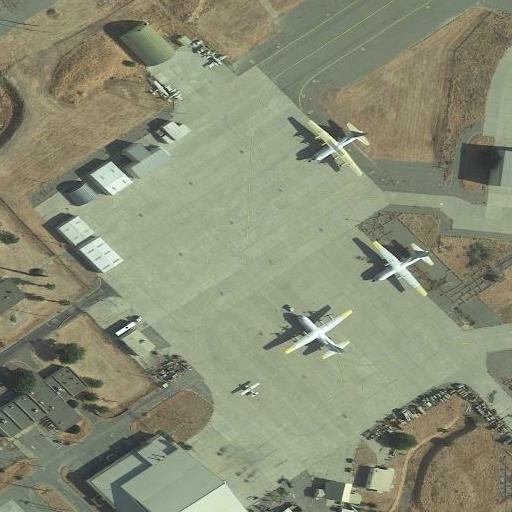}
\label{fig:le8}}
\subfigure[]{\includegraphics[width=0.3\textwidth]{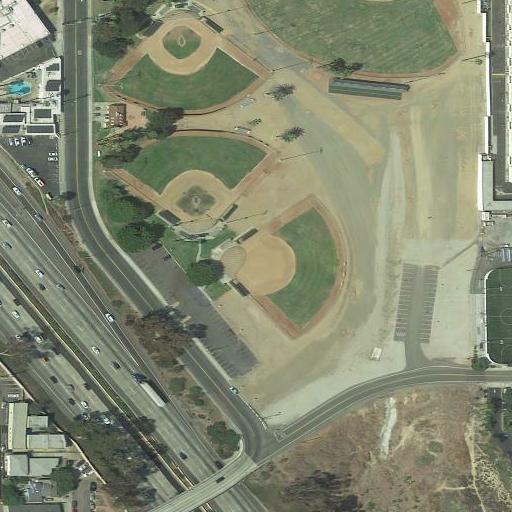}
\label{fig:le9}}
\vspace{-0.15cm}
\caption{Example images in our MAI dataset. Each image is $512 \times 512$ pixels, and their spatial resolutions range from 0.3 m/pixel to 0.6 m/pixel. We list their scene-level labels here: (a) farmland and residential; (b) baseball, woodland, parking lot, and tennis court; (c) commercial, parking lot, and residential; (d) woodland, residential, river, and runway; (e) river and storage tanks; (f) beach, woodland, residential, and sea; (g) farmland, woodland, and residential; (h) apron and runway; (i) baseball field, parking lot, residential, bridge, and soccer field.}
\label{fig:mai_example}
\end{figure*}

\begin{figure*}
    \centering
    \includegraphics[width=1\textwidth]{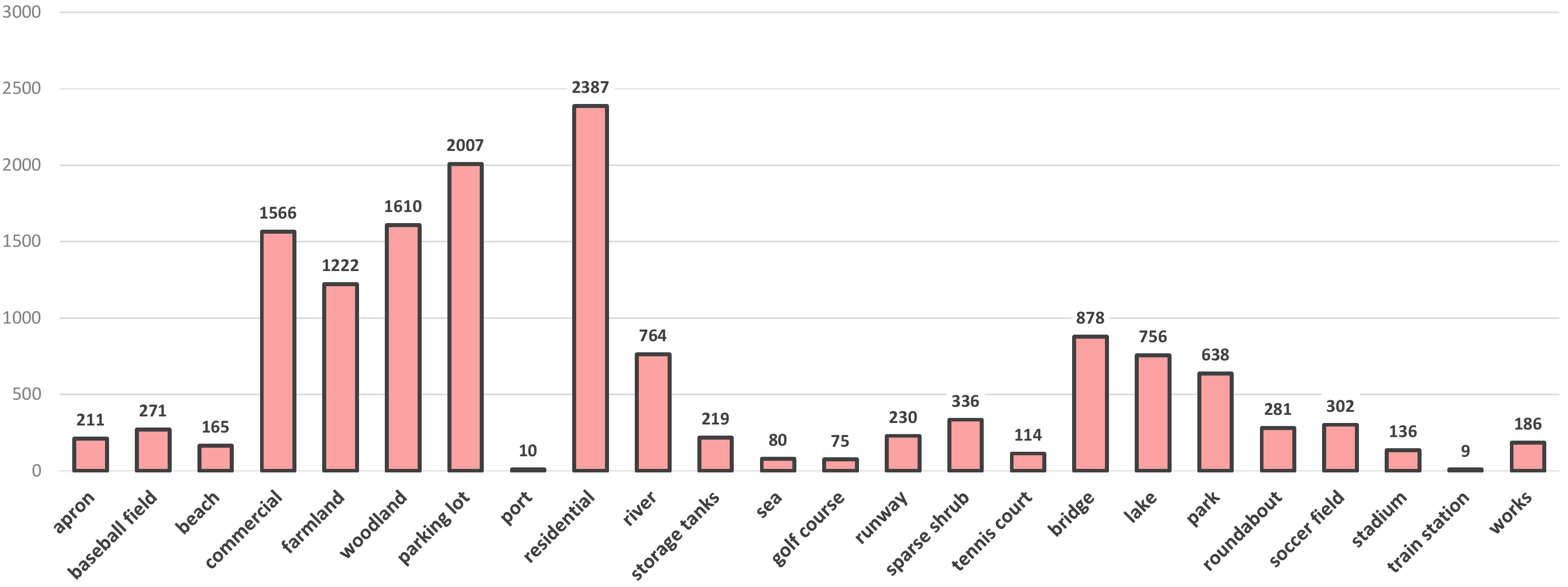}
    \caption{Statistics of the proposed MAI dataset for multi-scene classification in single aerial images.}
    \label{fig:mai_statistics}
\end{figure*}

\subsection{Dataset Description and Configuration}
\label{sec:data}
\subsubsection{MAI dataset}
To facilitate the progress of aerial scene interpretation in the wild, we yield a new dataset, MAI dataset, by collecting and labeling 3923 large-scale images from Google Earth imagery that covers the United States, Germany, and France. The size of each image is $512 \times 512$, and spatial resolutions vary from 0.3 m/pixel to 0.6 m/pixel. After capturing aerial images, we manually assign each image multiple scene-level labels from in total 24 scene categories, including apron, baseball, beach, commercial, farmland, woodland, parking lot, port, residential, river, storage tanks, sea, bridge, lake, park, roundabout, soccer field, stadium, train station, works, golf course, runway, sparse shrub, and tennis court. Notably, OSM data associated with the collected images cannot be directly employed as reference owing to the problems presented in Section~\ref{sec:intro}. Such a labeling procedure is extremely time- and labor-consuming, and annotating one image costs around 20 seconds, which is ten times more than labeling a single-scene image. Several example multi-scene images are shown in Figure~\ref{fig:mai_example}. Numbers of aerial images related to various scenes are reported in Figure~\ref{fig:mai_statistics}. Among existing datasets, BigEarthNet~\cite{sumbul2019bigearthnet} is one of the most relevant datasets, which consists of Sentinel-2 images acquired over the European Union with spatial resolutions ranging from 10 m/pixel to 60 m/pixel. Spatial sizes of images vary from $20 \times 20$ pixels to $120 \times 120$ pixels, and each is assigned multiple land-cover labels provided from the CORINE Land Cover map\footnote{https://land.copernicus.eu/pan-european/corine-land-cover}. Compared to BigEarthNet, our dataset is characterized by its high-resolution large-scale aerial images and worldwide coverage.

\subsubsection{UCM dataset}
\begin{figure}[!t]
\centering
\subfigure[]{\includegraphics[width=0.182\textwidth]{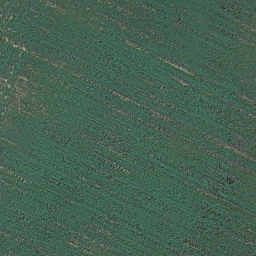}
\label{fig:ue1}}
\subfigure[]{\includegraphics[width=0.182\textwidth]{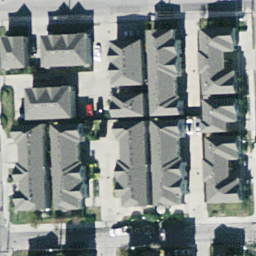}
\label{fig:ue2}}
\subfigure[]{\includegraphics[width=0.182\textwidth]{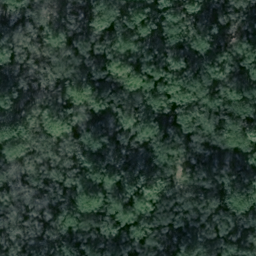}
\label{fig:ue3}}
\subfigure[]{\includegraphics[width=0.182\textwidth]{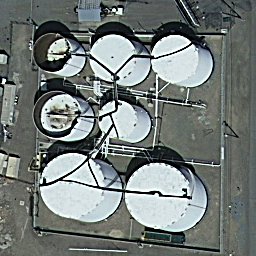}
\label{fig:ue4}}
\subfigure[]{\includegraphics[width=0.182\textwidth]{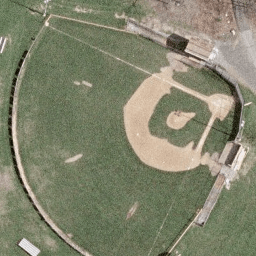}
\label{fig:ue9}}
\vspace{-0.5em}
\subfigure[]{\includegraphics[width=0.182\textwidth]{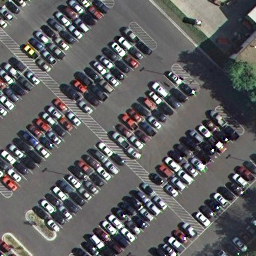}
\label{fig:ue5}}
\subfigure[]{\includegraphics[width=0.182\textwidth]{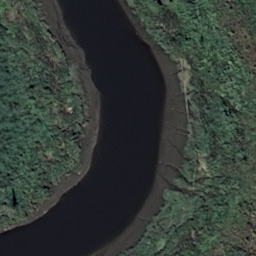}
\label{fig:ue6}}
\subfigure[]{\includegraphics[width=0.182\textwidth]{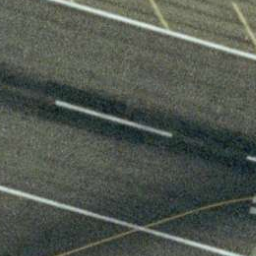}
\label{fig:ue7}}
\subfigure[]{\includegraphics[width=0.182\textwidth]{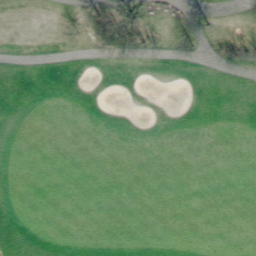}
\label{fig:ue8}}
\subfigure[]{\includegraphics[width=0.182\textwidth]{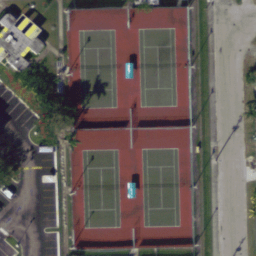}
\label{fig:ue10}}
\caption{Example single-scene aerial categories in the UCM dataset: (a) agricultural, (b) dense residential, (c) forest, (d) storage tanks, (e) baseball field, (f) parking lot, (g) river, (h) runway, (i) golf course, and (j) tennis court.}
\label{fig:ucm_example}
\end{figure}

UCM dataset~\cite{yang2010bag} is a commonly used single-scene aerial image dataset produced by Yang and Newsam from the University of California Merced. This dataset comprises 2100 aerial images cropped from aerial ortho imagery provided by the United States Geological Survey (USGS) National Map, and the spatial resolution of the collected images is one foot. The size of each image is $256 \times 256$ pixels, and all image samples are classified into 21 scene-level classes: overpass, forest, beach, baseball diamond, building, airplane, freeway, intersection, harbor, golf course, runway, agricultural, storage tank, mobile home park, medium residential, sparse residential, chaparral, river, tennis courts, dense residential, and parking lot. The number of aerial images collected for each scene is 100, and several example images are shown in Figure \ref{fig:ucm_example}. To learn scene prototypes from these single-scene images, we randomly choose 80\% of image samples per scene category to train and validate the embedding function and utilize the rest for testing.

\subsubsection{AID dataset}
AID dataset~\cite{xia2017aid} is a another popular single-scene aerial image dataset which consists of 10000 aerial images with a size of $600 \times 600$ pixels. These images are captured from Google Earth imagery that is taken over China, the United States, England, France, Italy, Japan, and Germany, and spatial resolutions of the collected images vary from 0.5 m/pixel to 8 m/pixel. In total, there are 30 scene categories, including viaduct, river, baseball field, center, farmland, railway station, meadow, bare land, storage tanks, beach, mountain, park, bridge, playground, church, commercial, desert, forest, parking, industrial, square, sparse residential, pond, medium residential, port, resort, airport, school, stadium, and dense residential. The number of images in different classes ranges from 220 to 420. Similar to the data split in the UCM dataset, 20\% of images are chosen from each scene as test samples, while the remaining images are utilized to train and validate the embedding function. Some example images of the AID dataset are exhibited in Figure~\ref{fig:aid_example}.

\subsubsection{Dataset configuration}
\label{sec:setup}
\begin{figure}[!t]
\centering
\subfigure[]{\includegraphics[width=0.182\textwidth]{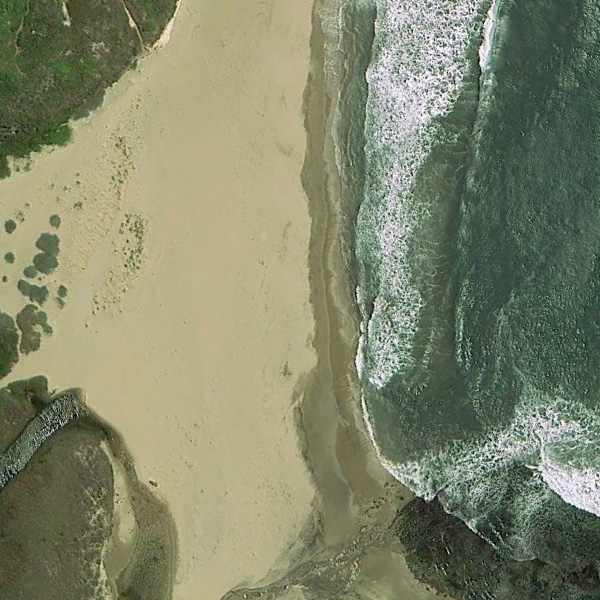}
\label{fig:ae1}}
\subfigure[]{\includegraphics[width=0.182\textwidth]{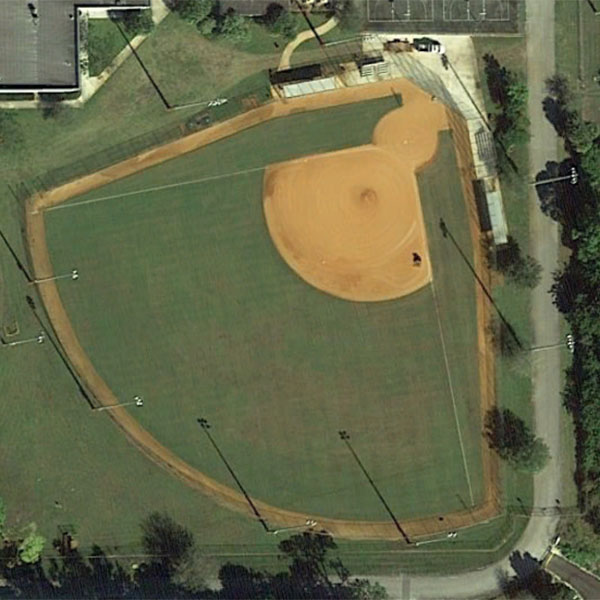}
\label{fig:ae2}}
\subfigure[]{\includegraphics[width=0.182\textwidth]{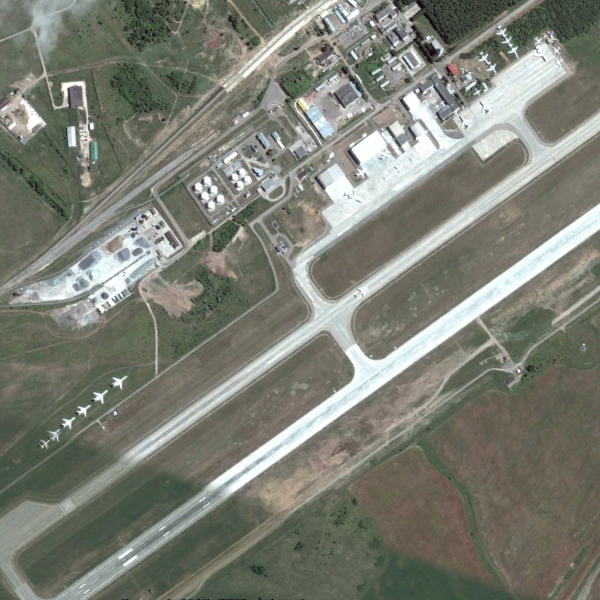}
\label{fig:ae3}}
\subfigure[]{\includegraphics[width=0.182\textwidth]{datasets/aid_example/railwaystation_8.jpg}
\label{fig:ae4}}
\subfigure[]{\includegraphics[width=0.182\textwidth]{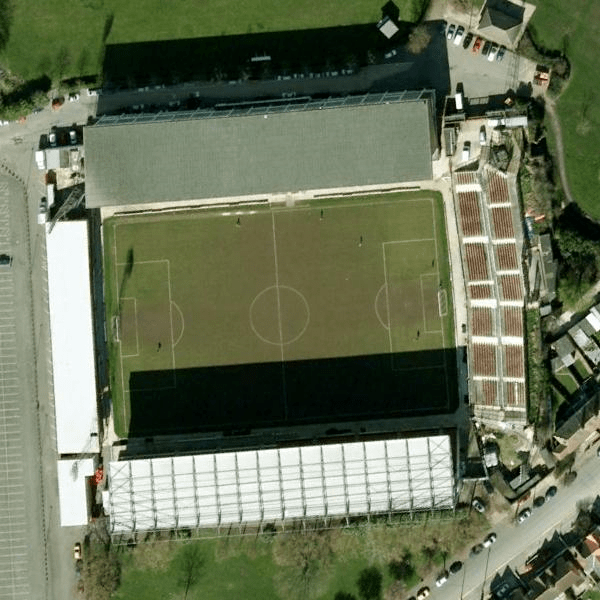}
\label{fig:ae5}}
\vspace{-0.5em}
\subfigure[]{\includegraphics[width=0.182\textwidth]{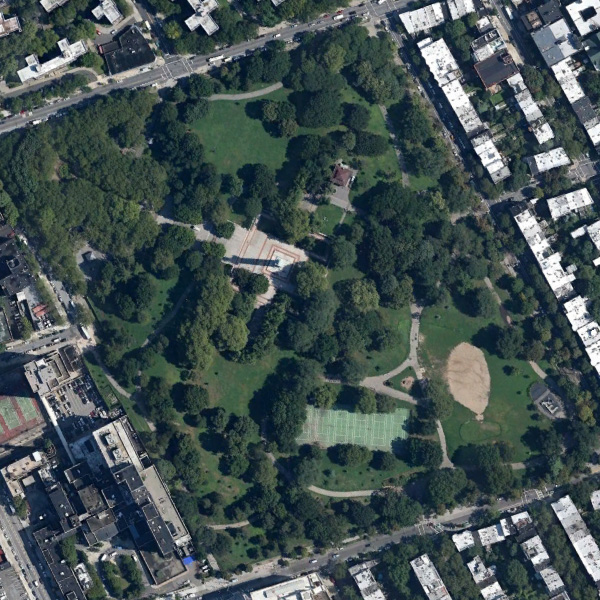}
\label{fig:ae6}}
\subfigure[]{\includegraphics[width=0.182\textwidth]{datasets/aid_example/playground_114.jpg}
\label{fig:ae7}}
\subfigure[]{\includegraphics[width=0.182\textwidth]{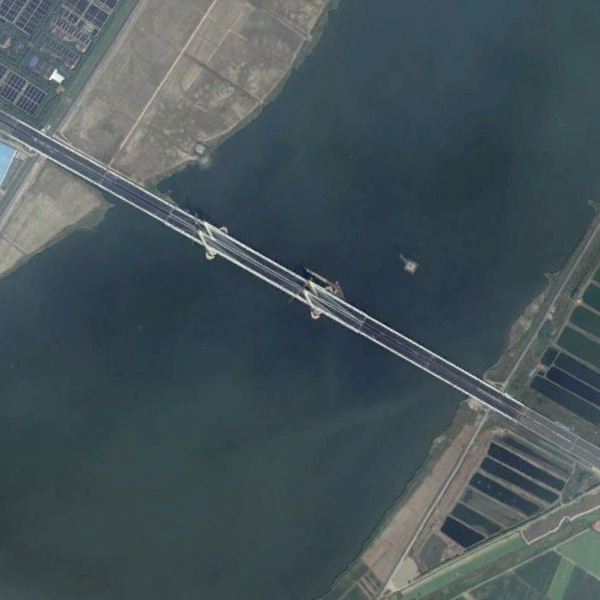}
\label{fig:ae8}}
\subfigure[]{\includegraphics[width=0.182\textwidth]{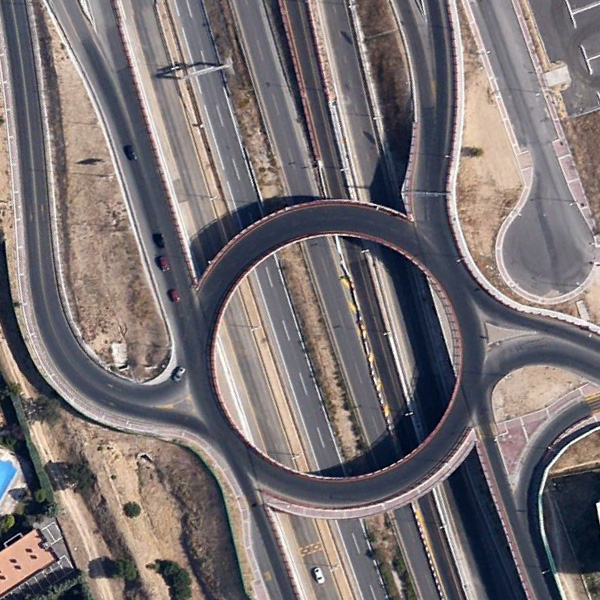}
\label{fig:ae9}}
\subfigure[]{\includegraphics[width=0.182\textwidth]{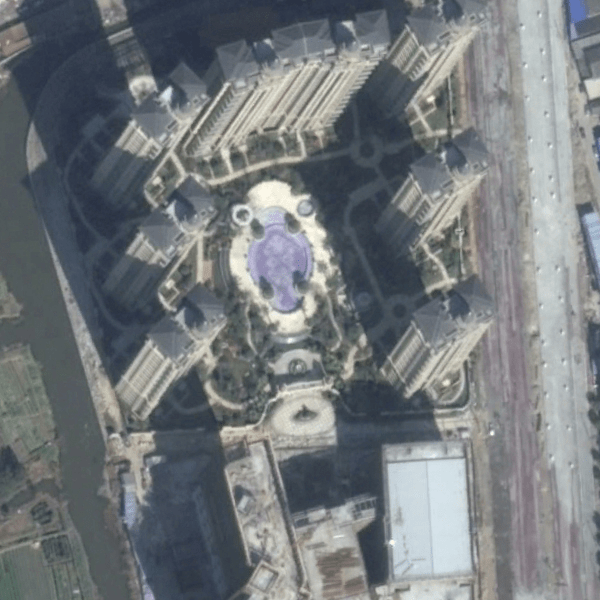}
\label{fig:ae10}}
\caption{Example single-scene aerial categories in the AID dataset: (a) beach, (b) baseball field, (c) airport, (d) railway station, (e) stadium, (f) park, (g) playground, (h) bridge, (i) viaduct, and (j) commercial.}
\label{fig:aid_example}
\end{figure}

In order to widely evaluate the performance of our method, we utilize two variant dataset configurations, UCM2MAI and AID2MAI, based on common scene categories shared by UCM/AID and MAI. Specifically, the UCM2MAI configuration consists of 1600 single-scene aerial images from the UCM dataset and 1649 multi-scene images from our MAI dataset. 16 aerial scenes that are commonly included in both two datasets are considered in UCM2MAI, and numbers of their associated images are listed in Table~\ref{tab:AID_UCM_MAI}. Besides, the AID2MAI configuration is composed of 7050 and 3239 aerial images from the AID and MAI datasets, respectively. 20 common scene categories are taken into consideration, and the number of images related to each scene is present in Table~\ref{tab:AID_UCM_MAI}. Although such configurations might limit the number of recognizable scene classes, we believe this limitation can be addressed by collecting more single-scene images by crawling OSM data and producing large-scale multi-scene aerial image datasets. We select only 90 and 120 multi-scene aerial images from UCM2MAI and AID2MAI as training instances, respectively, and test networks on the remaining multi-scene images. For rare scenes (e.g., port and train station), we select all associated training images, while for common scenes, we randomly select several of their training samples. It is noteworthy that we yield the scene prototype of \texttt{residential} by taking an average of high-level representations of aerial images belonging to scene \texttt{medium residential} and \texttt{dense residential}. Besides, although the UCM and AID datasets do not contain images for \texttt{sea}, their images for \texttt{beach} often comprise both sea and beach (cf. (c) in Figure~\ref{fig:aid_example}). Therefore, we make use of training samples labeled as \texttt{beach} to yield the prototype representation of \texttt{sea}.

\begin{table}[!t]
\centering
\renewcommand{\arraystretch}{1.1}
\caption{The Number of Images Associated with Each Scene.}
\label{tab:AID_UCM_MAI}
\begin{threeparttable}
\begin{tabular}{P{3.8cm}|P{1.5cm}P{1.5cm}|P{1.5cm}P{1.5cm}}
\Xhline{3\arrayrulewidth}
 & \multicolumn{2}{c|}{UCM2MAI} & \multicolumn{2}{c}{AID2MAI} \\
\hline
\hline
Scene Category & UCM & MAI & AID & MAI \\
\hline
apron & 100 & 194 & 360 & 54 \\
baseball field & 100 & 75 & 220 & 235 \\
beach & 100 & 94 & 400 & 130 \\
commercial & 100 & 607 & 350 & 1391 \\
farmland & 100 & 680 & 370 & 983 \\
woodland & 100 & 762 & 250 & 1312 \\
parking lot & 100 & 708 & 390 & 1777 \\
port & 100 & 3 & 380 & 9 \\
residential & 200 & 958 & 700 & 2082 \\
river & 100 & 209 & 410 & 686 \\
storage tanks & 100 & 89 & 360 & 193 \\
sea & 100* & 51 & 400* & 59 \\
golf course & 100 & 75 & - & - \\
runway & 100 & 230 & - & - \\
sparse shrub & 100 & 336 & - & - \\ 
tennis court & 100 & 114 & - & - \\
bridge & - & - & 360 & 878\\
lake & - & - & 420 & 756 \\
park & - & - & 350 & 638 \\
roundabout & - & - & 420 & 281 \\
soccer field & - & - & 370 & 302 \\
stadium & - & - & 290 & 136 \\
train station & - & - & 260 & 9 \\
works & - & - & 390 & 186 \\

\hline
\hline
All & 1600 & 1649 & 7050 & 3239 \\
\Xhline{3\arrayrulewidth}
\end{tabular}
\begin{tablenotes}
\item[] * indicates that the number of images is not counted in total amounts, as the scene prototype of \texttt{beach} and \texttt{sea} are learned from the same images.
\end{tablenotes}
\end{threeparttable}
\end{table}

\subsection{Training Details}
\label{sec:train}

The training procedure consists of two phases: 1) learning the embedding function $f_\phi$ on large quantities of single-scene aerial images and 2) training the entire PM-Net on a limited number of multi-scene images in an end-to-end manner. Thus, various training strategies are applied to each phase and detailed as follows. 

In the first training phase, the embedding function $f_\phi$ is initialized with the corresponding deep CNNs pretrained on ImageNet~\cite{imagenet_cvpr09}, and weights in $g_\theta$ are initialized by a Glorot uniform initializer. Eq.~(\ref{eq:cross_entropy}) is employed as the loss of the network, and Nestrov Adam \cite{nadam2} is chosen as the optimizer, of which parameters are set as recommended: $\beta_1 = 0.9$, $\beta_2 = 0.999$, and $\epsilon = 1e-08$. The learning rate is set as $2e-04$ and decayed by $\sqrt{0.1}$ when the validation loss fails to decrease for two epochs.

In the second learning phase, we initialize $f_\phi$ with parameters learned in the previous training stage and employ the Glorot uniform initializer to initialize all weights in ${\rm Q}_h$, ${\rm V}_h$, ${\rm K}_h$, and the last fully-connected layer. $L$ and $U$ are set to the same value of 256, and the number of heads is defined as 20. Notably, all weights are trainable, and the embedding function is tuned during the second training phase as well. Multiple scene-level labels are encoded as multi-hot vectors, where 0 indicates the absence of the corresponding scene while 1 refers to existing scenes. Accordingly, the loss is defined as binary cross-entropy. The optimizer is the same as that in the first training phase, but here we make use of a relatively large learning rate, $5e-4$. The network is implemented on TensorFlow and trained on one NVIDIA Tesla P100 16GB GPU for 100 epochs. We set the size of training batch to 32 for both training phases.

\begin{table}[!t]
\centering
\begin{adjustwidth}{-0.35cm}{0cm}
\renewcommand{\arraystretch}{1.3}
\caption{Differences between Two Training Phases.}
\label{tab:two-phase}
\begin{tabular}{c|c|c|c|c}
\Xhline{3\arrayrulewidth}
\multirow{2}{*}{Phase} & \multirow{2}{*}{Learnable Module} & \multicolumn{2}{c|}{Dataset} & \multirow{2}{*}{Memory} \\
\cline{3-4}
 & & Pretraining $f_\phi$ & Fine-tuning module &\\
\hline
\hline
1 & prototype learning & ImageNet & UCM/AID & updated \\
\hline
2 & memory retrieval & UCM/AID & MAI & frozen \\
\Xhline{3\arrayrulewidth}
\end{tabular}
\end{adjustwidth}
\end{table}

\subsection{Evaluation Metrics}

For the purpose of evaluating the performance of networks quantitatively, we utilize example-based $F_1$~\cite{wu2016unified} and $F_2$~\cite{f2} scores as evaluation metrics and calculate them with the following equation:
\begin{equation}
\label{eq:f2}
F_\beta = (1+\beta^2)\frac{p_er_e}{\beta^2p_e+r_e}, \hspace{1em} \beta=1,2,
\end{equation}
where $p_e$ and $r_e$ denote example-based precision and recall~\cite{tsoumakas2007random}. We calculate $p_e$ and $r_e$ as follows:
\begin{equation}
\label{eq:pere}
p_e = \frac{TP_e}{TP_e+FP_e},\hspace{1em} r_e = \frac{TP_e}{TP_e+FN_e},
\end{equation}
where $FN_e$, $FP_e$, and $TP_e$ represent numbers of false negatives, false positives, and true positives in an example, respectively. In our case, an example is a multi-scene aerial image, and by averaging scores of all examples in the test set, the mean example-based $F$ scores, precision, and recall can be eventually computed. In addition to example-based evaluation metrics, we also calculate label-based precision $p_l$ and recall $r_l$ with Eq.~\ref{eq:pere} but replace $FN_e$, $FP_e$, and $TP_e$ with numbers of false negatives, false positives, and true positives in respect of each scene category. The mean $p_l$ and $r_l$ can then be calculated. Note that principle indexes are the mean $F_1$ and $F_2$ scores.

\begin{table}[!t]
\centering
\renewcommand{\arraystretch}{1.3}
\caption{Numerical Results on UCM2MAI (\%).}
\label{tab:ucm2MAI}
\begin{adjustwidth}{-0.55cm}{0cm}
\begin{threeparttable}
\begin{tabular}{ccccccc}
\Xhline{3\arrayrulewidth}
Model & m. $F_1$ & m. $F_2$ & m. $p_e$ & m. $r_e$ & m. $p_l$ & m. $r_l$ \\
\hline
VGGNet*~\cite{simonyan2014very} & 32.16 & 32.79 & 35.08 & 34.35 & 21.74 & 22.57 \\
VGGNet~\cite{simonyan2014very} & 51.42 & 49.04 & 62.00 & 48.38 & 36.80 & \textbf{27.44} \\
Mem-N2N-VGGNet~\cite{sukhbaatar2015end} & 52.16 & 50.93 & 57.26 & \textbf{50.73} & 20.79 & 22.58 \\
K-Branch CNN~\cite{sumbul2019novel} & 47.04 & 43.15 & 64.57 & 41.83 & 37.93 & 22.28 \\
proposed PM-VGGNet & \textbf{54.42} & \textbf{51.16} & \textbf{67.35} & 49.95 & \textbf{47.24} & 26.79 \\
\hline
Inception-V3*~\cite{szegedy2015going} & 48.03 & 44.37 & 62.22 & 42.80 & 47.36 & 20.43 \\
Inception-V3~\cite{szegedy2015going} & 53.96 & 51.28 & \textbf{65.47} & 50.49 & \textbf{51.03} & \textbf{32.88} \\
Mem-N2N-Inception-V3~\cite{sukhbaatar2015end} & 56.06 & 55.27 & 62.95 & 55.92 & 47.90 & 30.48 \\
proposed PM-Inception-V3 & \textbf{58.56} & \textbf{58.06} & 64.17 & \textbf{58.73} & 46.44 & 26.47\\
\hline
ResNet*~\cite{he2016deep} & 48.36 & 45.00 & 63.90 & 43.84 & 53.63 & 28.35 \\
ResNet~\cite{he2016deep} & 51.39 & 48.31 & 65.33 & 47.37 & 51.89 & \textbf{30.54} \\
Mem-N2N-ResNet~\cite{sukhbaatar2015end} & 54.31 & 51.45 & 63.97 & 50.31 & 44.33 & 24.58 \\
proposed PM-ResNet & \textbf{56.89} & \textbf{54.11} & \textbf{69.85} & \textbf{53.38} & \textbf{55.93} & 29.76 \\
\hline
NASNet*~\cite{nas} & 43.64 & 39.94 & 58.56 & 38.39 & 46.01 & 19.69 \\
NASNet~\cite{nas} & 52.03 & 49.43 & 64.24 & 48.75 & 49.99 & \textbf{33.75} \\
Mem-N2N-NASNet~\cite{sukhbaatar2015end} & 55.17 & 53.05 & 64.71 & 52.65 & 49.60 & 29.14 \\
proposed PM-NASNet & \textbf{60.13} & \textbf{59.57} & \textbf{67.04} & \textbf{60.42} & \textbf{58.60} & \textbf{35.04} \\
\Xhline{3\arrayrulewidth}
\end{tabular}
\begin{tablenotes}
\item[]CNN* is initialized with weights pretrained on ImageNet.
\item[]CNN, Mem-N2N, and PM-Net are initialized with parameters pretrained on the UCM dataset.
\item[]m.$F_1$ and m.$F_2$ indicate the mean $F_1$ and $F_2$ score.
\item[]m. $p_e$ and m. $r_e$ indicate mean example-based precision and recall.
\item[]m. $p_l$ and m. $r_l$ indicate mean label-based precision and recall.
\end{tablenotes}
\end{threeparttable}
\end{adjustwidth}
\end{table}

\begin{table}[!t]
\caption{Example Images and Predictions on UCM2MAI.}
\label{tab:predictions_ucm2MAI}
\centering
\begin{adjustwidth}{-1.75cm}{0cm}
\renewcommand{\arraystretch}{1.3}
\begin{threeparttable}
\begin{tabular}{>{\centering\arraybackslash}P{2.9cm}>{\centering\arraybackslash}m{2.9cm}>{\centering\arraybackslash}m{2.9cm}>{\centering\arraybackslash}m{2.9cm}>{\centering\arraybackslash}m{2.9cm}} 
\Xhline{3\arrayrulewidth}
Sample Multi-scene Aerial Images from MAI Dataset
& \subfigure{\includegraphics[width=0.2\textwidth]{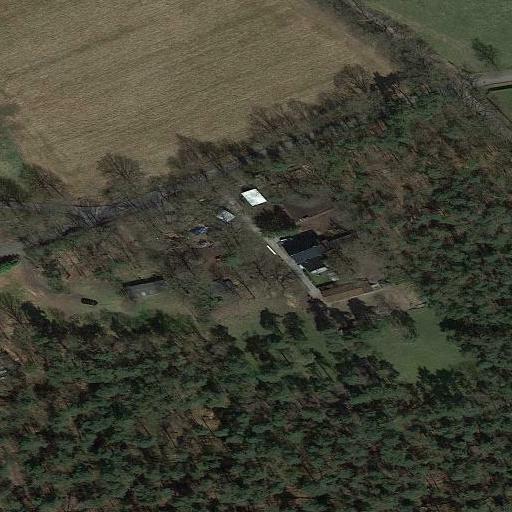}}
& \subfigure{\includegraphics[width=0.2\textwidth]{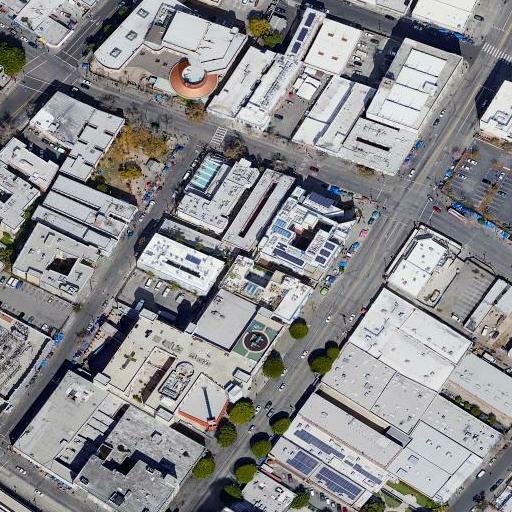}} 
& \subfigure{\includegraphics[width=0.2\textwidth]{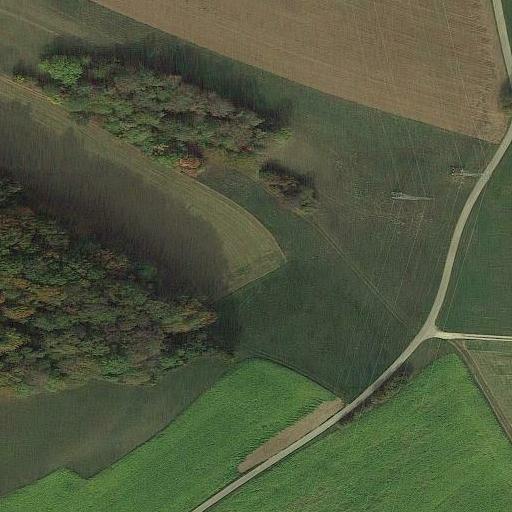}} 
& \subfigure{\includegraphics[width=0.2\textwidth]{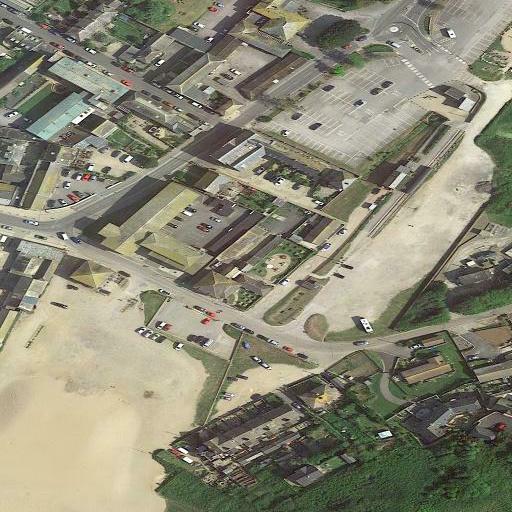}}\\
\hline
Ground Truths & farmland, woodland, residential & commercial, parking lot, residential & woodland, farmland & commercial, beach, parking lot, residential \\
\hline
Predictions &  farmland, woodland, residential & commercial, parking lot, residential & woodland, farmland & commercial, beach, parking lot, residential \\
\hline
\hline
Sample Multi-scene Aerial Images from MAI Dataset
& \subfigure{\includegraphics[width=0.2\textwidth]{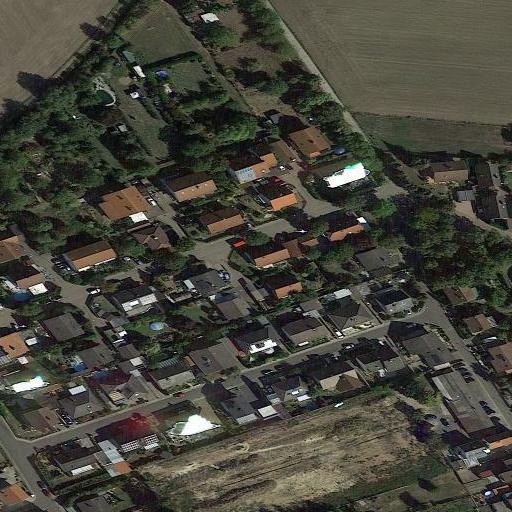}}
& \subfigure{\includegraphics[width=0.2\textwidth]{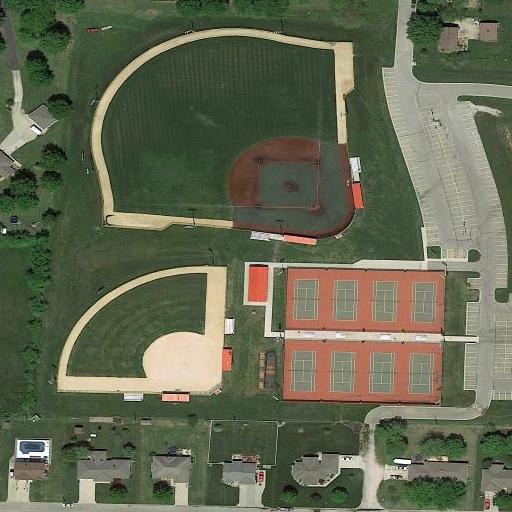}} 
& \subfigure{\includegraphics[width=0.2\textwidth]{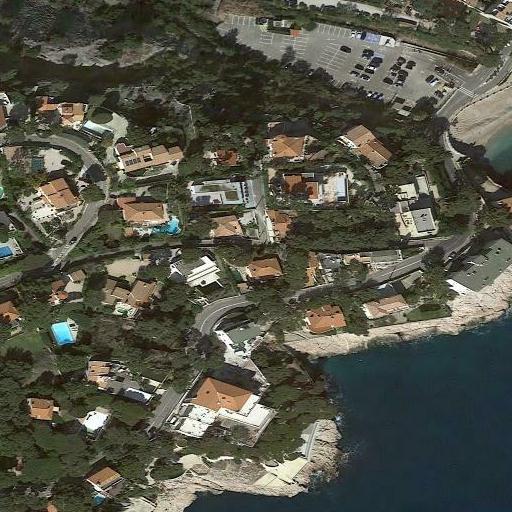}}
& \subfigure{\includegraphics[width=0.2\textwidth]{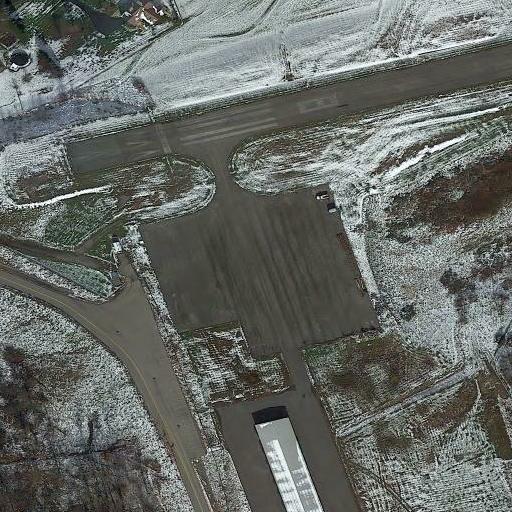}}\\
\hline
Ground Truths & farmland, parking lot, residential & baseball field, parking lot, residential, tennis court & beach, parking lot, woodland, residential, sea & apron, runway\\
\hline
Predictions & \textcolor{blue}{farmland}, parking lot, residential & baseball field, parking lot, residential, \textcolor{blue}{tennis court} & \textcolor{red}{commercial}, \textcolor{blue}{beach}, parking lot, \textcolor{blue}{woodland}, residential, \textcolor{blue}{sea} & \textcolor{blue}{apron}, \textcolor{red}{residential}, \textcolor{blue}{runway}, \textcolor{red}{parking lot}\\
\Xhline{3\arrayrulewidth}
\end{tabular}
\begin{tablenotes}
\item[]\textcolor{blue}{Blue} predictions are false negatives, while \textcolor{red}{red} predictions indicate false positives.
\end{tablenotes}
\end{threeparttable}
\end{adjustwidth}
\end{table}

\begin{table}[!t]
\centering
\renewcommand{\arraystretch}{1.3}
\caption{Numerical results on AID2MAI (\%).}
\label{tab:aid2MAI}
\begin{adjustwidth}{-0.35cm}{0cm}
\begin{threeparttable}
\begin{tabular}{ccccccc}
\Xhline{3\arrayrulewidth}
Model & m. $F_1$ & m. $F_2$ & m. $p_e$ & m. $r_e$ & m. $p_l$ & m. $r_l$ \\
\hline
VGGNet*~\cite{simonyan2014very} & 41.57 & 36.36 & 64.02 & 34.04 & 25.98 & 12.80 \\
VGGNet~\cite{simonyan2014very} & 48.30 & 50.80 & 48.53 & \textbf{54.19} & 32.89 & \textbf{44.75} \\
Mem-N2N-VGGNet~\cite{sukhbaatar2015end} &  45.92 & 43.17 & 56.16 & 42.22 & 23.10 & 18.76 \\
K-Branch CNN~\cite{sumbul2019novel} & 47.67 & 43.88 & 63.84 & 42.37 & 26.53 & 16.15 \\
proposed PM-VGGNet & \textbf{54.37} & \textbf{51.44} & \textbf{65.69} & 50.39 & \textbf{48.06} & 22.40 \\
\hline
Inception-V3*~\cite{szegedy2015going} & 45.92 & 40.76 & 66.17 & 38.43 & 39.56 & 14.71 \\
Inception-V3~\cite{szegedy2015going} & 51.81 & 49.44 & 62.91 & 48.93 & 45.26 & \textbf{36.32} \\
Mem-N2N-Inception-V3~\cite{sukhbaatar2015end} & 52.13 & \textbf{53.83} & 52.53 & \textbf{56.21} & 33.33 & 29.05 \\
proposed PM-Inception-V3 & \textbf{53.08} & 49.26 & \textbf{69.42} & 47.85 & \textbf{48.20} & 24.65 \\
\hline
ResNet*~\cite{he2016deep} & 50.06 & 46.88 & 64.32 & 45.98 & 39.48 & 22.34 \\
ResNet~\cite{he2016deep} & 54.74 & 52.76 & 65.54 & 52.62 & 47.54 & \textbf{40.23} \\
Mem-N2N-ResNet~\cite{sukhbaatar2015end} & 53.26 & \textbf{60.41} & 46.15 & \textbf{68.07} & 23.75 & 30.21 \\
proposed PM-ResNet & \textbf{57.42} & 54.34 & \textbf{70.62} & 53.33 & \textbf{55.34} & 29.55\\
\hline
NASNet*~\cite{nas} & 47.53 & 42.93 & \textbf{65.57} & 40.94 & 34.79 & 16.42 \\
NASNet~\cite{nas} & 53.08 & 50.68 & 64.33 & 50.17 & \textbf{46.68} & \textbf{37.43} \\
Mem-N2N-NASNet~\cite{sukhbaatar2015end} & 39.27 & 40.72 & 38.52 & 42.38 & 20.03 & 20.41 \\
proposed PM-NASNet & \textbf{54.11} & \textbf{52.39} & 64.03 & \textbf{52.30} & 43.16 & 33.99 \\
\Xhline{3\arrayrulewidth}
\end{tabular}
\begin{tablenotes}
\item[]CNN, Mem-N2N, and PM-Net are initialized with parameters pretrained on the AID dataset.
\end{tablenotes}
\end{threeparttable}
\end{adjustwidth}
\end{table}

\begin{table}
\caption{Example Images and Predictions on AID2MAI.}
\label{tab:predictions_aid2MAI}
\centering
\begin{adjustwidth}{-1.75cm}{0cm}
\renewcommand{\arraystretch}{1.3}
\begin{tabular}{>{\centering\arraybackslash}P{2.9cm}>{\centering\arraybackslash}m{2.9cm}>{\centering\arraybackslash}m{2.9cm}>{\centering\arraybackslash}m{2.9cm}>{\centering\arraybackslash}m{2.9cm}} 
\Xhline{3\arrayrulewidth}
Sample multi-scene aerial images from MAI dataset  & 
\subfigure{\includegraphics[width=0.2\textwidth]{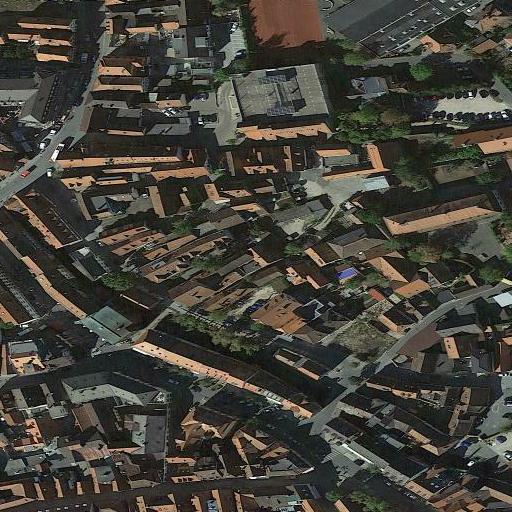}} &
\subfigure{\includegraphics[width=0.2\textwidth]{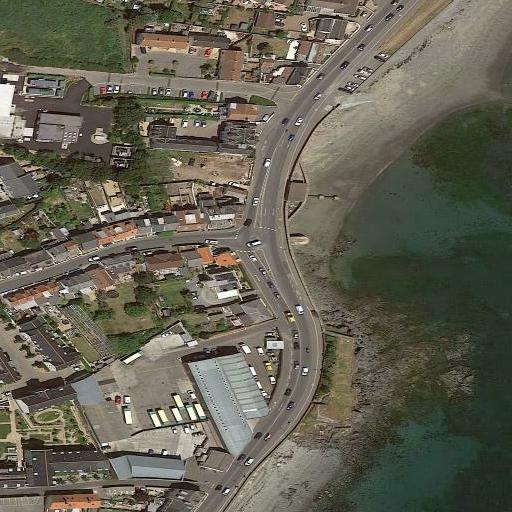}} & 
\subfigure{\includegraphics[width=0.2\textwidth]{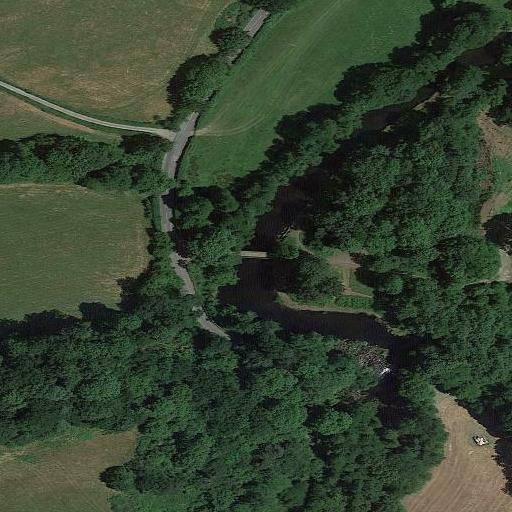}} & 
\subfigure{\includegraphics[width=0.2\textwidth]{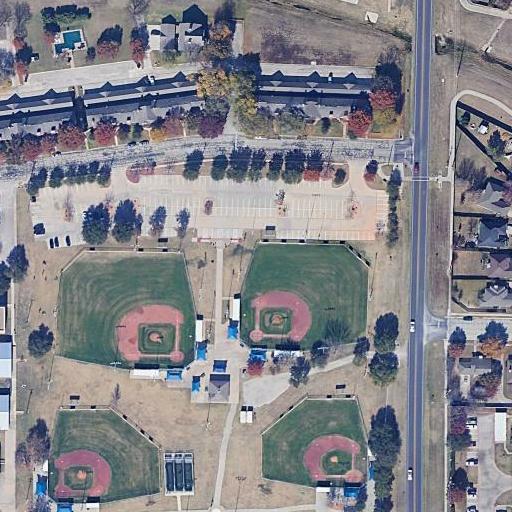}}
\\
\hline
Ground Truths & bridge, river, commercial, parking lot, residential & beach, commercial, parking lot, residential, sea & bridge, farmland, river, woodland & baseball field, parking lot, park, residential\\
\hline
Predictions & bridge, river, commercial, parking lot, residential & beach, commercial, parking lot, residential, sea & bridge, farmland, river, woodland & baseball field, parking lot, park, residential \\
\hline
Sample multi-scene aerial images from MAI dataset  
& \subfigure{\includegraphics[width=0.2\textwidth]{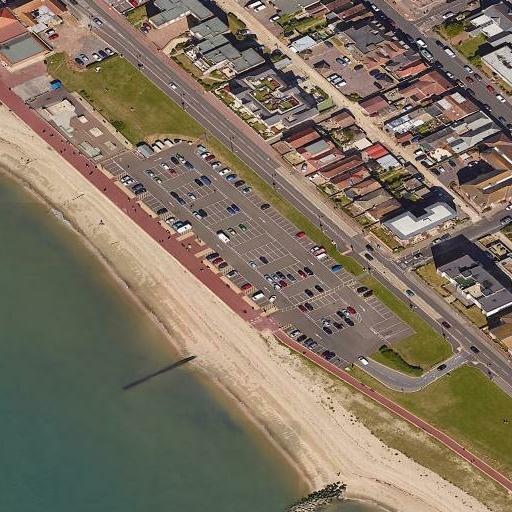}}
& \subfigure{\includegraphics[width=0.2\textwidth]{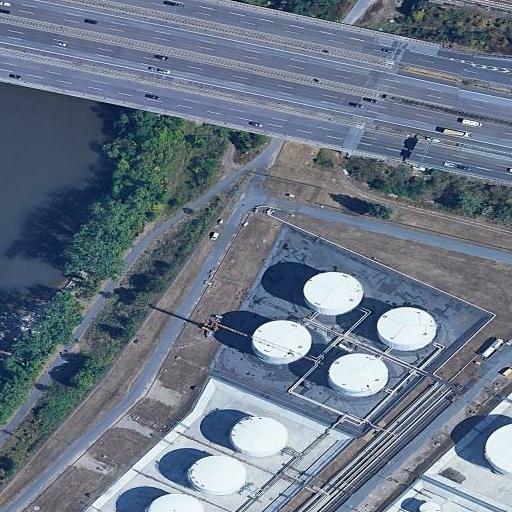}} 
& \subfigure{\includegraphics[width=0.2\textwidth]{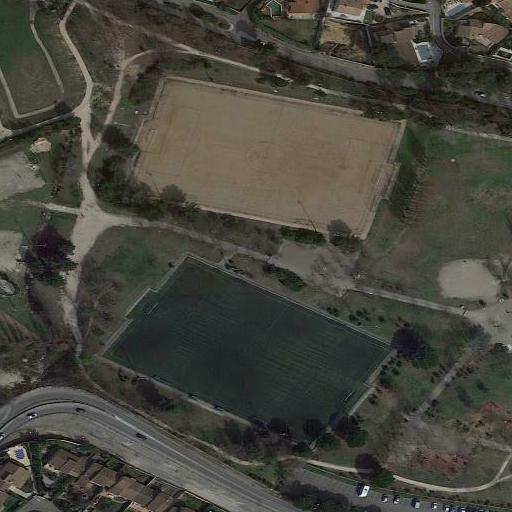}}
& \subfigure{\includegraphics[width=0.2\textwidth]{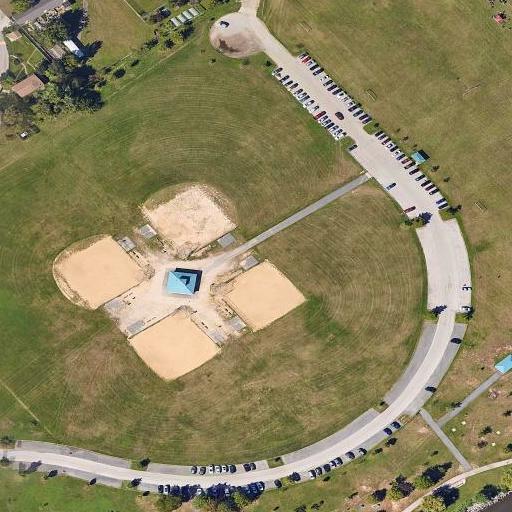}}\\
\hline
Ground Truths & beach, commercial, parking lot, residential, sea & bridge, woodland, river, storage tanks & baseball field, commercial, parking lot, park, residential, soccer field & baseball field, parking lot, soccer field\\
\hline
Predictions & beach, commercial, parking lot, residential, sea, \textcolor{red}{woodland} & bridge, woodland, \textcolor{red}{farmland}, \textcolor{blue}{river}, \textcolor{blue}{storage tanks}, \textcolor{red}{parking lot} & \textcolor{blue}{baseball field}, \textcolor{blue}{commercial}, \textcolor{red}{woodland}, parking lot, \textcolor{blue}{park}, \textcolor{blue}{soccer field}, \textcolor{blue}{residential} & baseball field, \textcolor{red}{commercial}, \textcolor{red}{woodland}, parking lot, \textcolor{red}{park}, \textcolor{blue}{soccer field}, \textcolor{red}{residential}\\
\Xhline{3\arrayrulewidth}
\end{tabular}
\end{adjustwidth}
\end{table}

\subsection{Results on UCM2MAI}
\label{sec:out_aid2MAI}

For a comprehensive evaluation, we compare the proposed PM-Net with two baselines, CNN* and CNN. The former is initialized with parameters pretrained on ImageNet, and the latter is pretrained on single-scene datasets. Besides, we compare our network with a memory network, Mem-N2N~\cite{sukhbaatar2015end}. Since Mem-N2N was proposed for the question answering task, we adapt it to our task by replacing its inputs, i.e., embeddings of \textit{questions} and \textit{statements}, with \textit{query image representations} $f_\phi(\bm{X})$ and \textit{scene prototypes} $\bm{p}_s$, respectively. To be more specific, we feed $\bm{X}$ to a CNN backbone and take its output as the input of Mem-N2N. Scene prototypes are stored in the memory of Mem-N2N and retrieved according to $f_\phi(\bm{X})$. The initialization of $f_\phi$ is the same as that of our network, and the entire Mem-N2N is trained in an end-to-end manner. Various backbones of embedding functions are test, and quantitative results are reported in Table \ref{tab:ucm2MAI}. Besides, we also compare with a multi-attention driven multi-label classification network, termed as K-Branch CNN~\cite{sumbul2019novel}. K-Branch samples images into $K$ spatial resolutions and extracts their features with separate branches. Afterwards, a bidirectional recurrent neural network is employed to encode their relationships for inferring multiple labels. In our experiments, $K$ is set as default, 3, and input sizes of the three branches are $224 \times 224$, $112 \times 112$, and $56 \times 56$, respectively. Here we analyze results from the following three perspectives.

\subsubsection{The effectiveness of learnt single-scene prototypes}
To demonstrate the effectiveness of the prototype-inhabiting external memory, here we focus on comparisons between PM-Net and standard CNNs. In Table~\ref{tab:ucm2MAI}, PM-VGGNet increases the mean $F_1$ and $F_2$ scores by $3.00\%$ and $2.12\%$, respectively, with respect to VGGNet, and PM-ResNet obtains increments of $5.50\%$ and $5.80\%$ in the mean $F_1$ and $F_2$ scores compared to ResNet. Besides, it is interesting to observe that PM-NASNet achieves not only the best mean $F_1$ and $F_2$ scores ($60.13\%$ and $59.57\%$) but also relatively high example-based precision and recall in comparison with other competitors. This demonstrates that employing NASNet as the embedding function can enhance the robustness of PM-Net. Comparisons between PM-Inception-V3 with Inception-V3 show that the external memory module contributes to improvements of $4.60\%$ and $6.78\%$ in the mean $F_1$ and $F_2$ scores, respectively. To summarize, memorizing and leveraging scene prototypes learned from huge quantities of single-scene images can improve the performance of network in multi-label scene recognition when limited training samples are available. For a deep insight, we further conduct ablation studies on the prototype modality and embedding function.

\begin{figure}[!t]
\centering
\includegraphics[width=.8\textwidth]{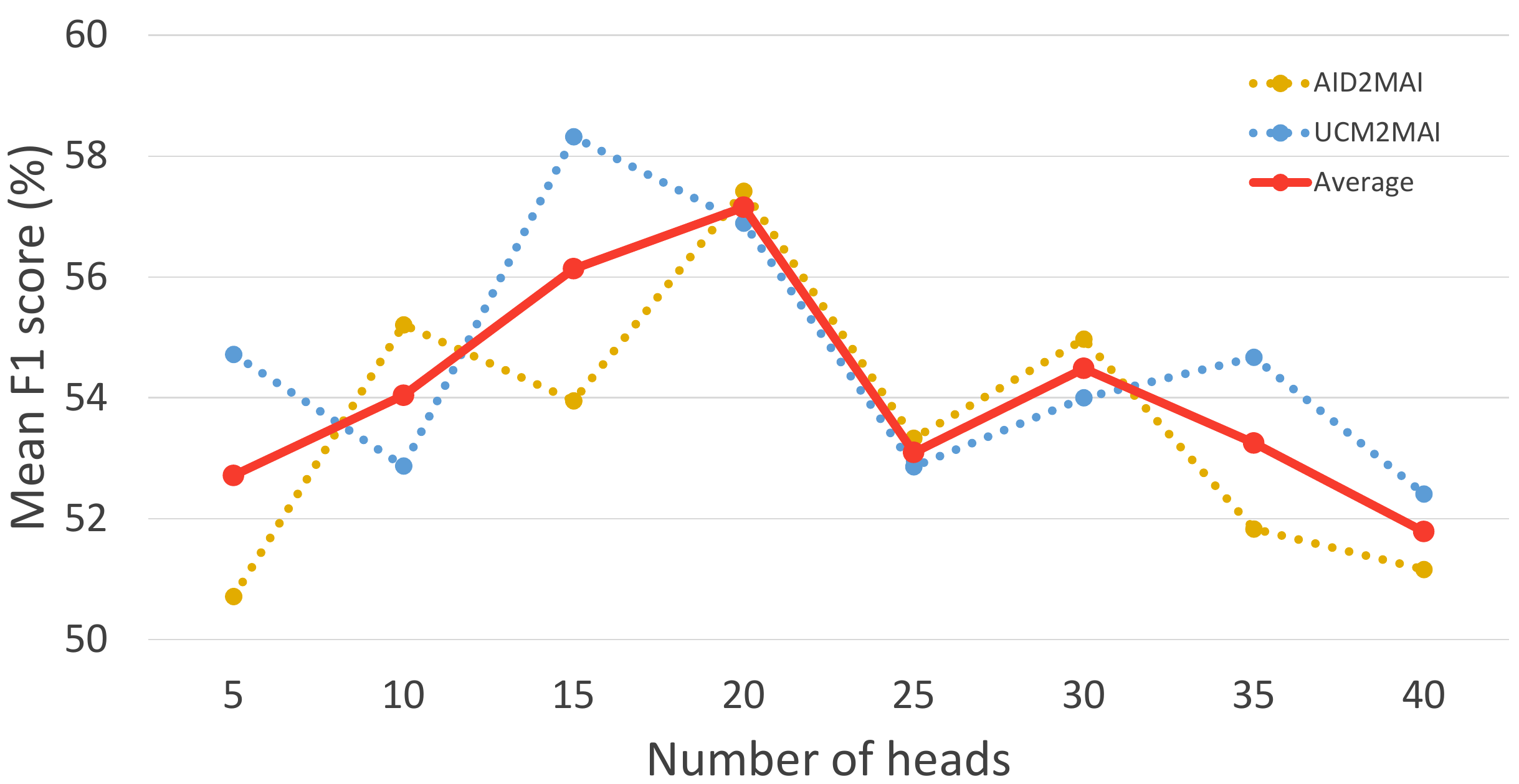}
\caption{The influence of the number of heads on both dataset configurations. \textcolor{Lblue}{Blue} and \textcolor{Lyellow1}{yellow} dot lines represent mean $F_1$ scores on UCM2MAI and AID2MAI. The \textcolor{Lred}{Red} line indicates the average of them.}
\label{fig:head_exp}
\end{figure}

\begin{figure}[!t]
\centering
\includegraphics[width=.8\textwidth]{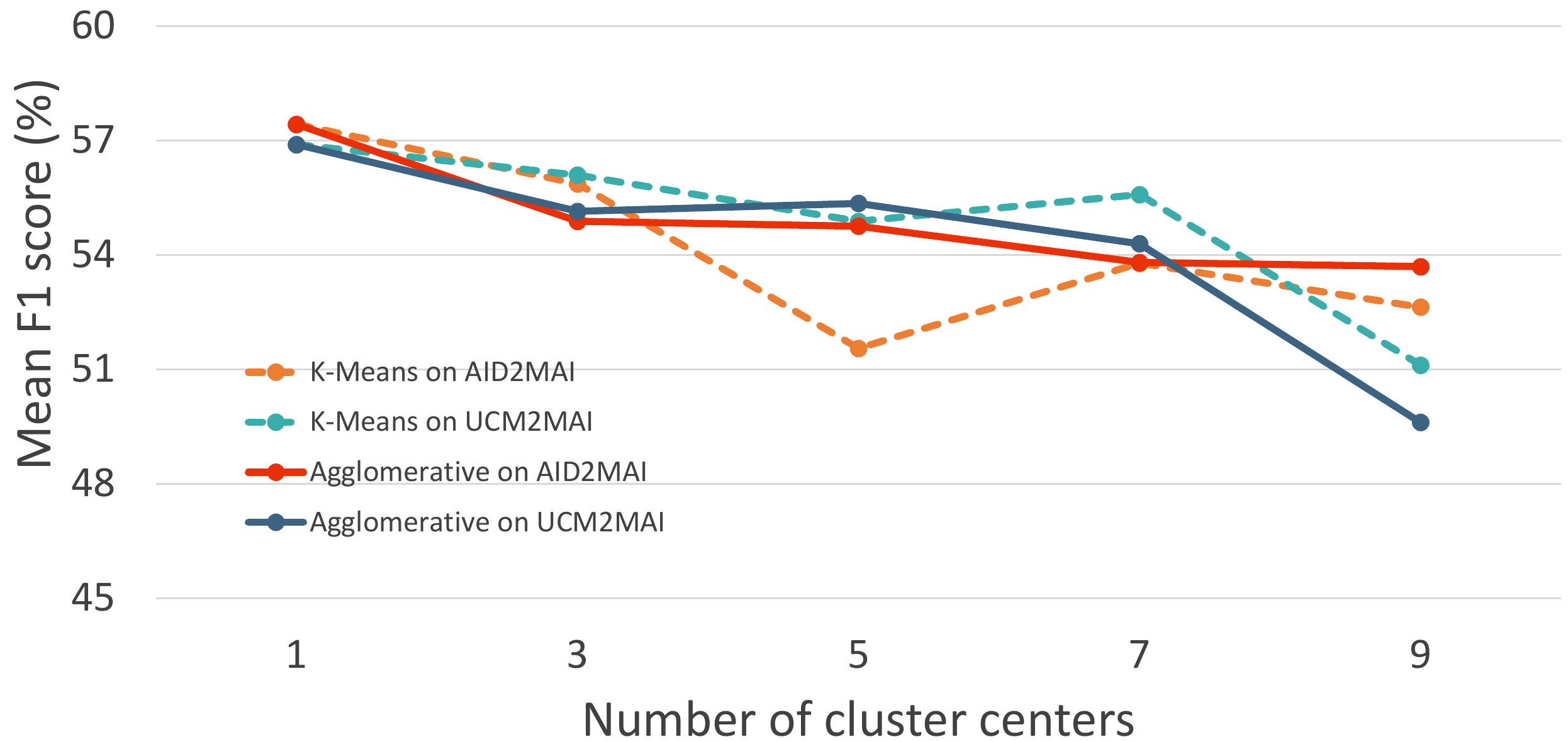}
\caption{The influence of the number of cluster centers on both dataset configurations. K-Means (\textcolor{Dblue2}{turquoise} and \textcolor{Dred2}{orange} dash lines) and Agglomerative (\textcolor{Dblue1}{blue} and \textcolor{Dred1}{red} lines) clustering algorithms are tested with PM-ResNet on both UCM2MAI and AID2MAI, respectively.}
\label{fig:cluster_exp}
\end{figure}

\textbf{Single- vs. multi-prototype representations.} We note that images collected over variant countries show high intra-class variability, and therefore, we wonder whether learning multi-prototype scene representations could improve the effectiveness of PM-Net. Specifically, instead of yielding scene prototypes via Eq.~\ref{eq:mean_ps}, we partition representations of single-scene aerial images belonging to the same scene into several clusters and take cluster centers as multi-prototype representations of each scene. In our experiments, we test two clustering methods, K-Means~\cite{kmeans} and Agglomerative~\cite{agglomerative}, with PM-ResNet on both UCM2MAI and AID2MAI, and results are shown in Figure~\ref{fig:cluster_exp}. We can see that the performance of PM-ResNet is decreased with the increasing number of cluster centers either using K-Means or Agglomerative clustering algorithms. Explanations could be that there are no obvious subclusters within each scene category (cf. Figure~\ref{fig:vis_tsne}), and thus PM-Net does not benefit from fine-grained multi-prototype representations.

\begin{figure}[!t]
\centering
\subfigure[]{\includegraphics[width=.49\textwidth]{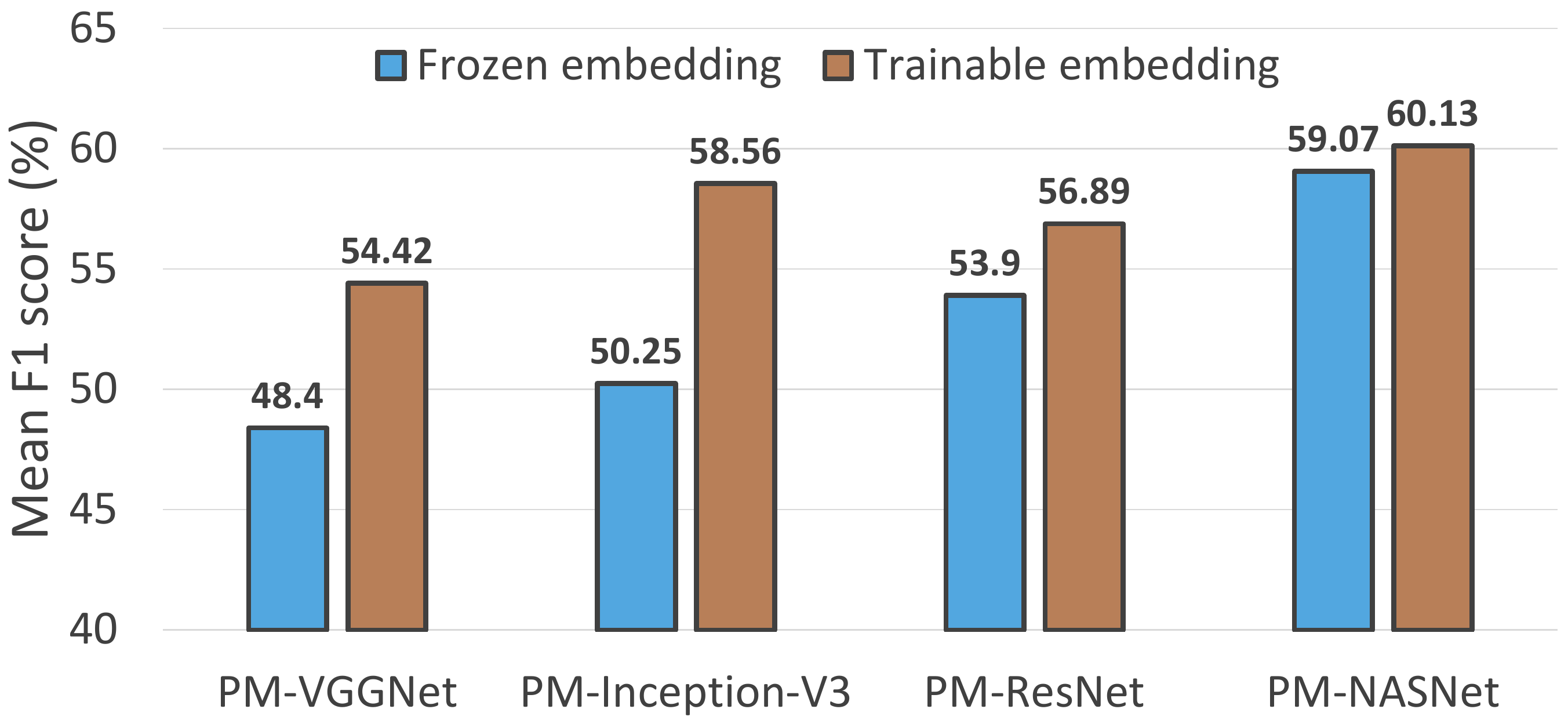}}
\subfigure[]{\includegraphics[width=.49\textwidth]{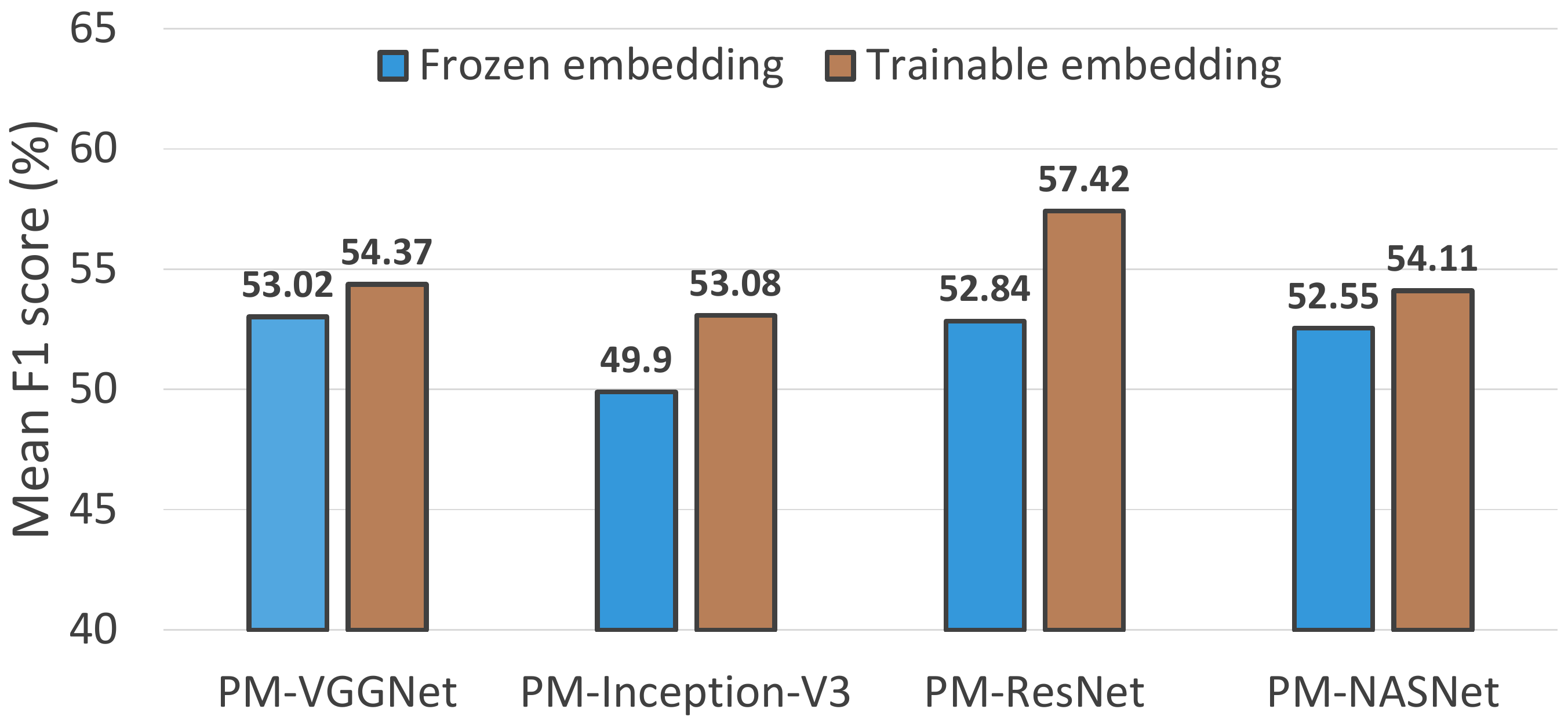}}
\caption{Comparisons between freezing and fine-tuning embedding functions on (a) UCM2MAI and (b) AID2MAI, respectively. \textcolor{Lblue1}{Blue} bars represent the performance of PM-Net with frozen embedding functions, and \textcolor{brown}{brown} bars denote the performance of PM-Net with trainable embedding functions.}
\label{fig:frozen_embed}
\end{figure}

\textbf{Frozen vs. trainable embedding function.} The embedding function plays a key role in both scene prototype learning and memory retrieval. In the former, we train the embedding function on single-scene images, while in the latter, the function is fine-tuned on multi-scene images. To explore the effectiveness of fine-tuning, we conduct experiments on freezing the embedding function when learning the memory retrieval module. The comparisons between PM-Net learned with frozen and trainable embedding functions are shown in Figure~\ref{fig:frozen_embed}. It can be observed that PM-Net with a trainable embedding function shows higher performance on both UCM2MAI and AID2MAI configurations. The reason could be that sources of single- and multi-scene images are variant, and fine-tuning can narrow their gaps.

\begin{figure}[!t]
\centering
\subfigure[]{\includegraphics[width=.49\textwidth]{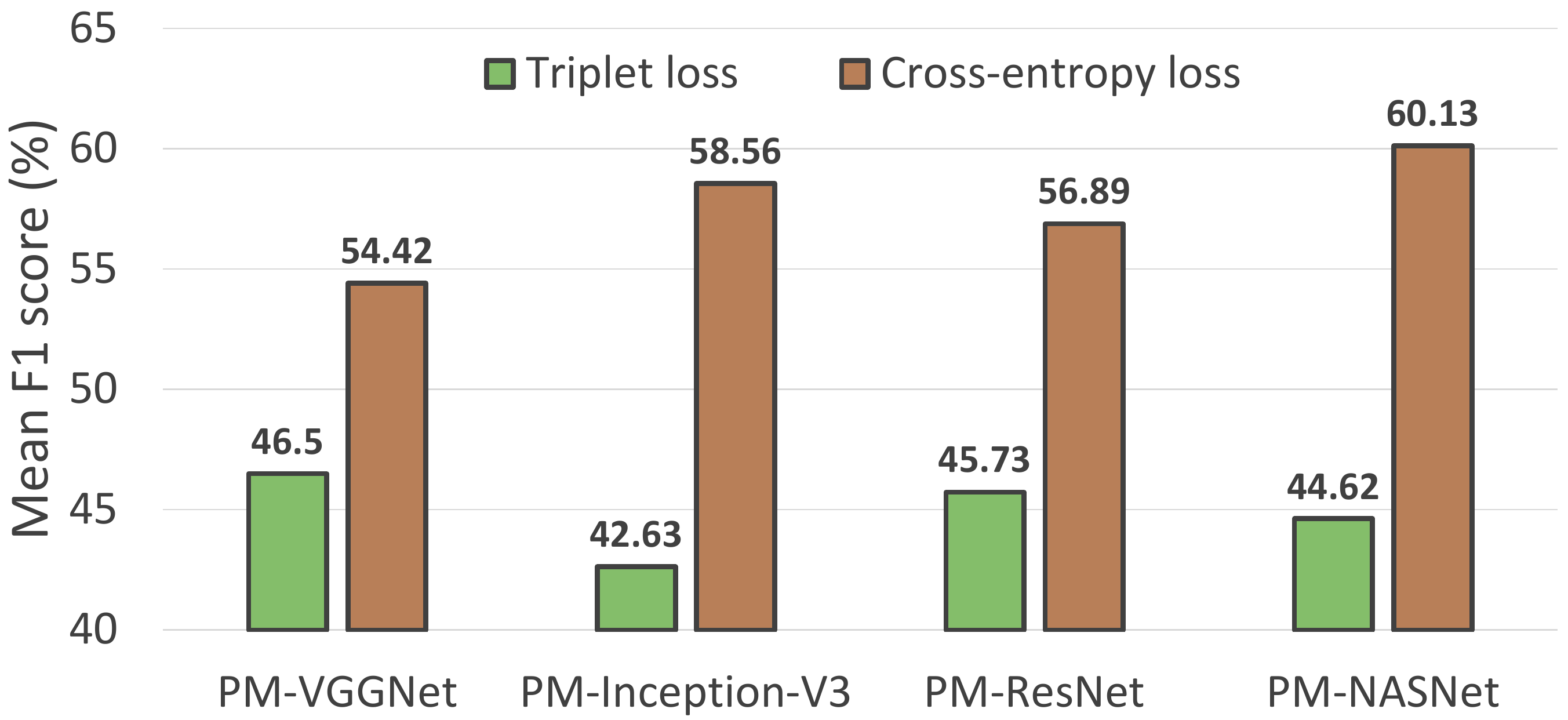}}
\subfigure[]{\includegraphics[width=.49\textwidth]{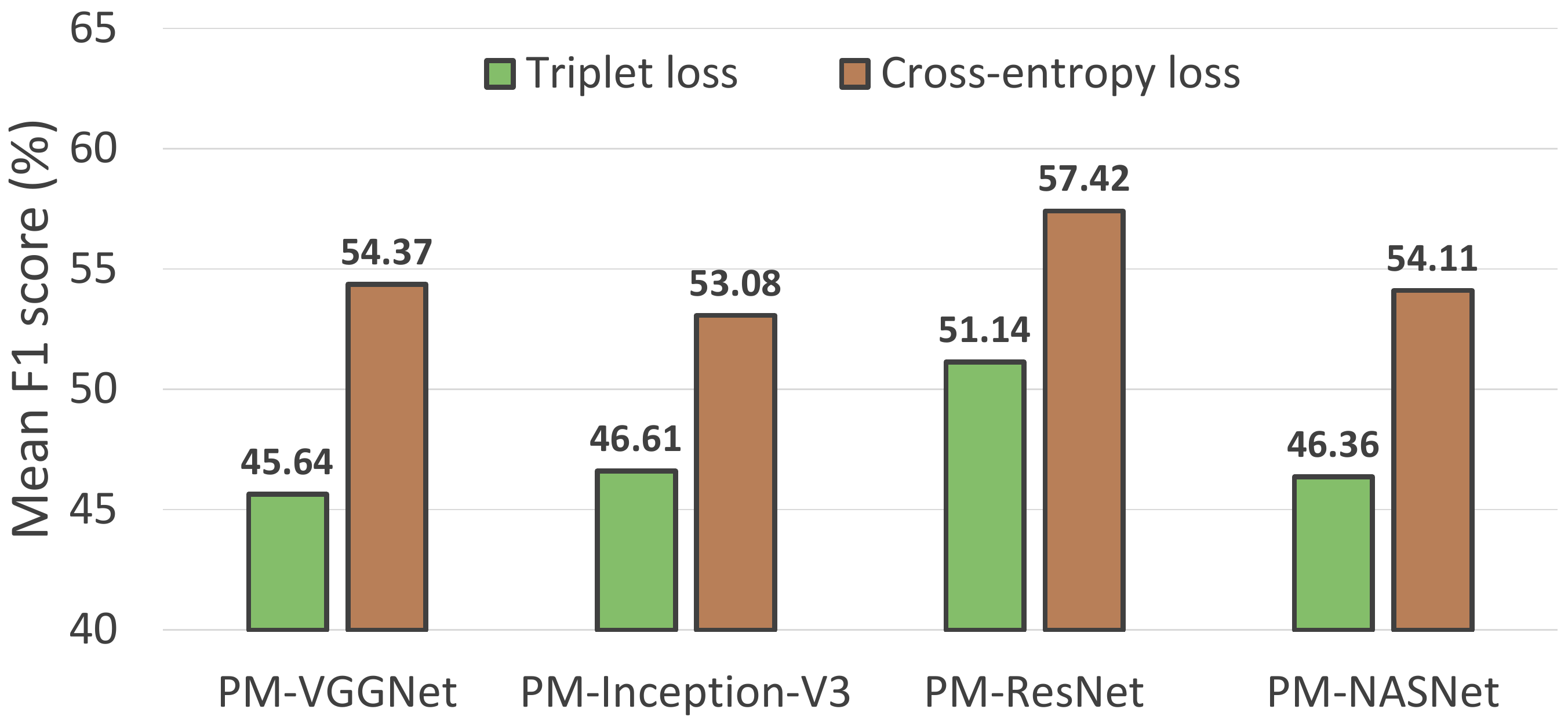}}
\caption{Comparisons of different loss functions on (a) UCM2MAI and (b) AID2MAI, respectively. \textcolor{Lgreen}{Green} bars denote the performance of PM-Net using embedding functions trained by the triplet loss, and \textcolor{brown}{brown} bars denote the performance of PM-Net with the cross-entropy loss as $\mathcal{L}$.}
\label{fig:triplet_loss}
\end{figure}

\textbf{Triplet vs. cross-entropy loss.} Triplet loss~\cite{schroff2015facenet} is known as learning discriminative representations by minimizing distances between embeddings of the same class while pushing away those of different classes. To study its performance in our task, we train the embedding function by replacing Eq.~\ref{eq:cross_entropy} with the following equation:
\begin{equation}
\label{eq:triplet}
    \mathcal{L}(\bm{X}^s_{i}) = max(||f_\phi(\bm{X}^s_{i})-f_\phi(\bm{X}^s_{pos})||^2-||f_\phi(\bm{X}^s_{i})-f_\phi(\bm{X}^s_{neg})||^2+\alpha, 0),
\end{equation}
where $\bm{X}^s_{pos}$ and $\bm{X}^s_{neg}$ denote positive and negative samples, i.e., images belonging to common and different classes, respectively, and $\alpha$ is set as default, $0.5$. The trained embedding function is then utilized to extract scene prototypes and initialize $f_\phi$ in the phase of learning the memory retrieval module. Besides, all the other setups are remained the same. We compare the performance of PM-Net using embedding functions trained through different loss functions in Figure~\ref{fig:triplet_loss}. It can be seen that training embedding functions with the triplet loss leads to decrements of the network performance. This can be attributed to that limited numbers of positive and negative samples in each batch can lead to local optimum. More specifically, the size of training batches is 32, and the number of scenes are 16 and 20 in UCM2MAI and AID2MAI, respectively. Thus, it is high probably that only a certain number of scenes are included in one batch, and comprehensively modeling relations between embeddings of samples from all scenes is infeasible. This also illustrates the larger performance decay on UCM2MAI compared to AID2MAI.

\subsubsection{The effectiveness of our multi-head attention-based memory retrieval module} As a key component of the proposed PM-Net, the multi-head attention-based memory retrieval module is designed to retrieve scene prototypes from the external memory, and we evaluate its effectiveness by comparing PM-Net with Mem-N2N. As shown in Table~\ref{tab:ucm2MAI}, PM-Net outperforms Mem-N2N with variant embedding functions. Specifically, PM-VGGNet increases the mean $F_1$ and $F_2$ scores by $2.26\%$ and $0.23\%$, respectively, compared to Mem-N2N-VGGNet. While taking ResNet as the embedding function, the improvement can reach $2.58\%$ in the mean $F_1$ score. Besides, the highest increments of mean $F_1$ and $F_2$ scores, $4.96\%$ and $6.52$, are achieved by PM-NASNet. These observations demonstrate that our memory retrieval module plays a key role in inferring multiple aerial scenes. An explanation could be that compared to the memory reader in Mem-N2N, our module comprise multiple heads, and each of them focuses on encoding a specific relevance between the query image and variant scene prototypes. In this case, more comprehensive scene-related memories can be used for inferring multiple scene labels. Moreover, we analyze the influence of the number of heads in the memory retrieval module. Figure~\ref{fig:head_exp} shows mean $F_1$ scores achieved by PM-Net with variant head numbers on both UCM2MAI and AID2MAI. We can observe that the network performance is first boosted with an increasing number of heads and then decreased gradually when the number exceeds 20. 

\begin{figure}[!t]
\centering
\subfigure[]{\includegraphics[width=.49\textwidth]{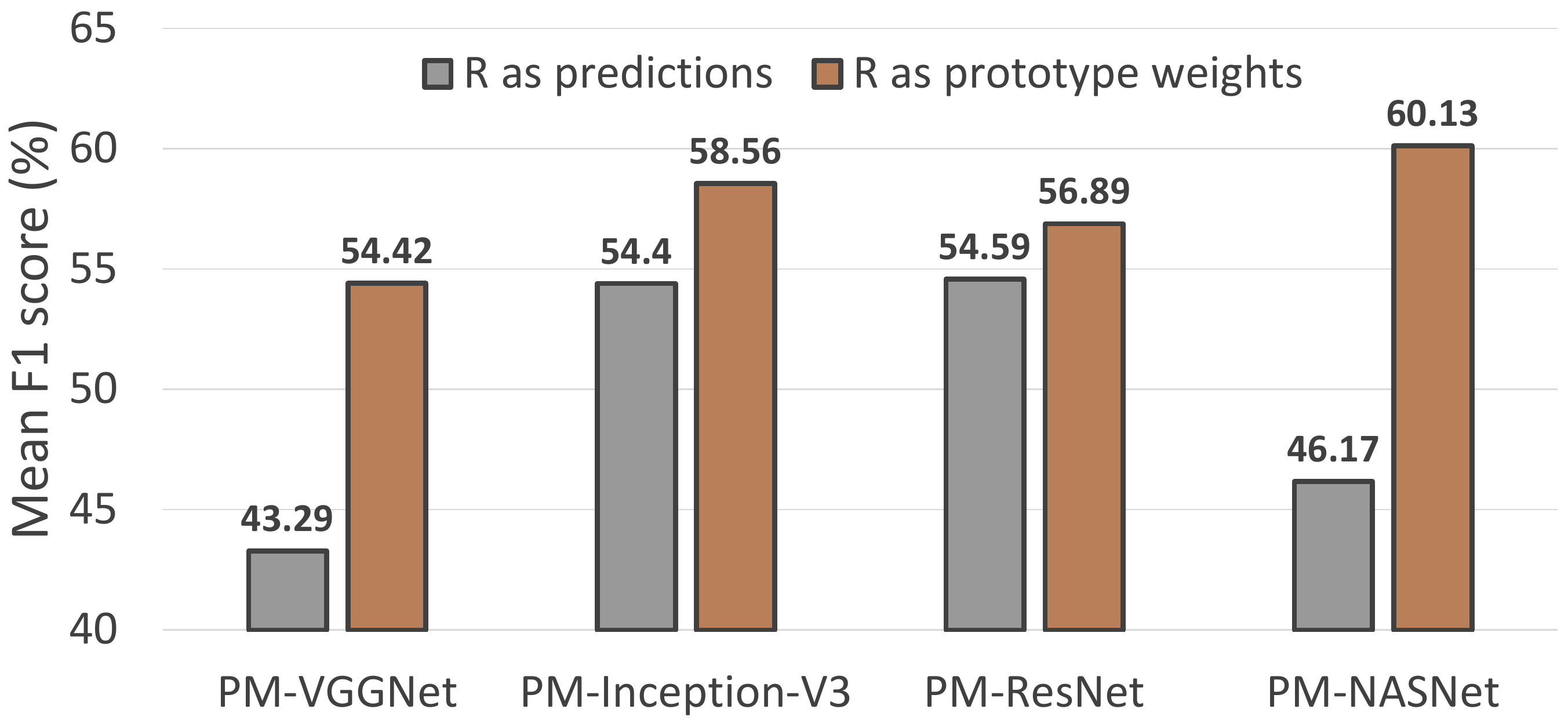}}
\subfigure[]{\includegraphics[width=.49\textwidth]{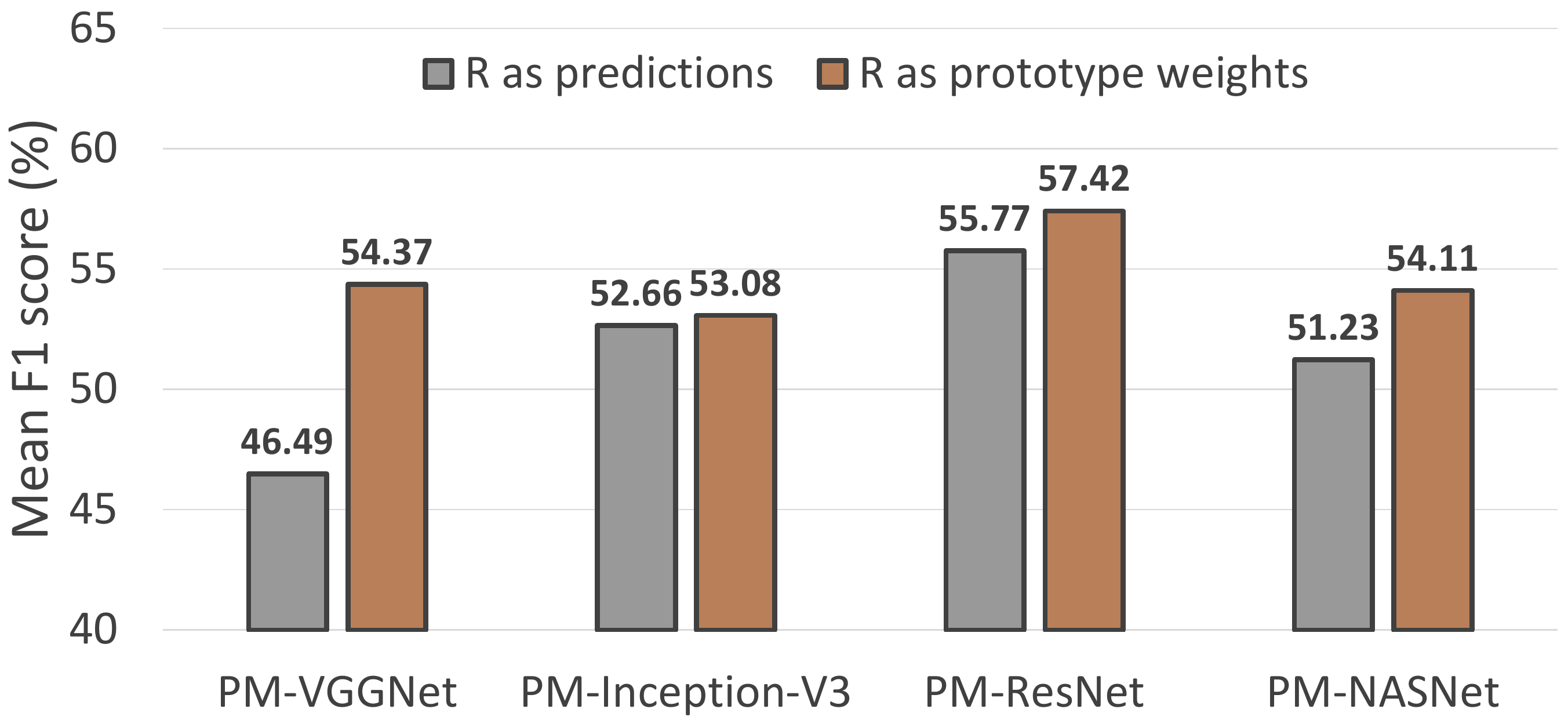}}
\caption{Comparisons between taking relevance ${\rm R}(\bm{X}, \bm{M})$ as predictions and prototype weights on (a) UCM2MAI and (b) AID2MAI, respectively. \textcolor{Lgray}{Gray} and \textcolor{brown}{brown} bars represent the performance of PM-Net making predictions from relevances and aggregated scene prototypes, respectively.}
\label{fig:r}
\end{figure}

Moreover, we also conduct experiments on directly utilizing relevances for inferring multiple scene labels. Specifically, we set the number of heads to 1 and replace the softmax activation in Eq.~\ref{eq:attention} with the sigmoid function. Relevances between the query image and scene prototypes can then be interpreted as the existence of each scene. We compare it with our memory retrieval module on variant backbones, and results are shown in Figure~\ref{fig:r}. We can see that utilizing relevances ${\rm R}(\bm{X}, \bm{M})$ as weights for aggregating scene prototypes leads to higher network performance.


\subsubsection{The benefit of exploiting single-scene training samples}
\begin{figure}[!t]
\centering
\subfigure[]{\includegraphics[width=.49\textwidth]{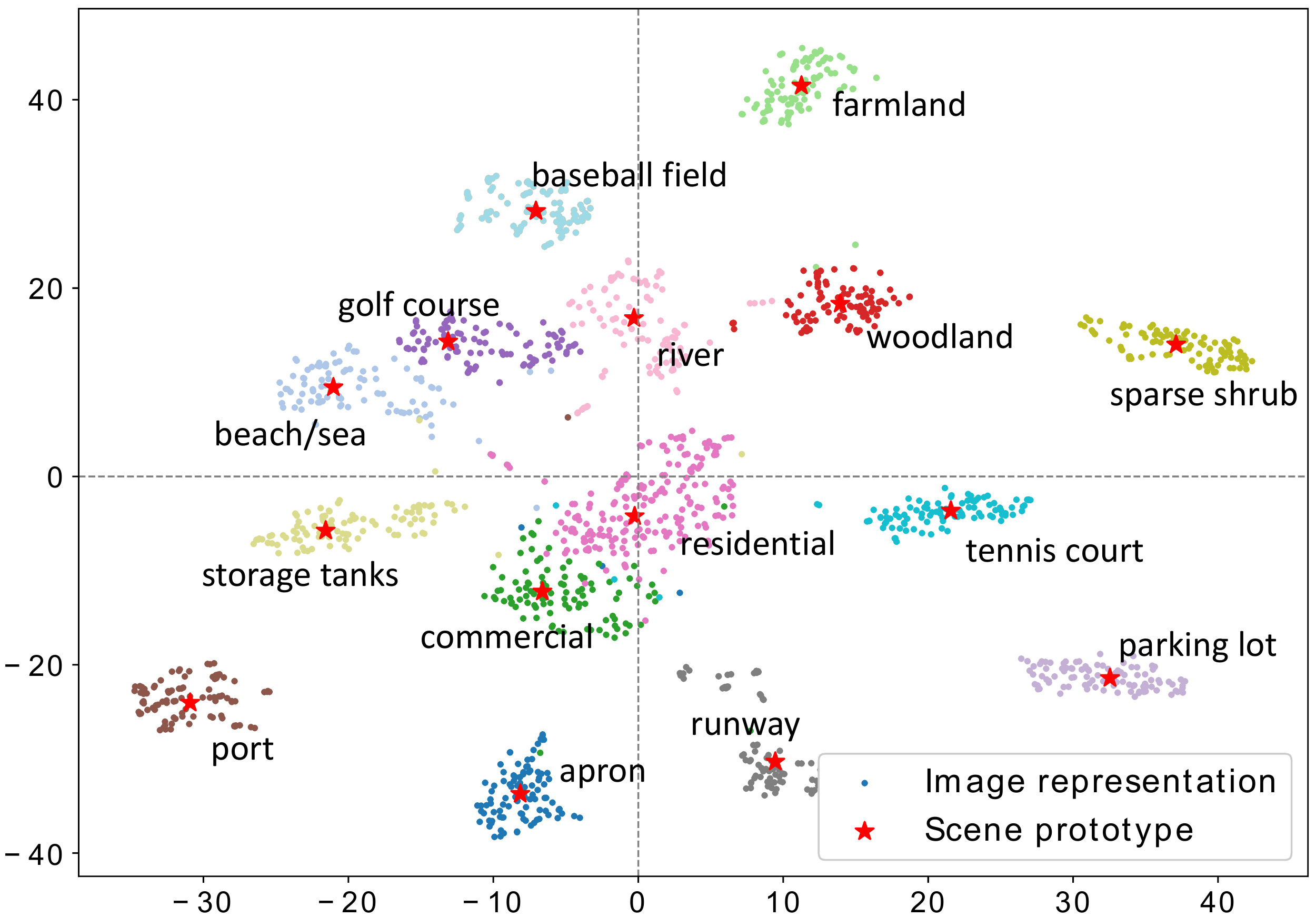}}
\subfigure[]{\includegraphics[width=.49\textwidth]{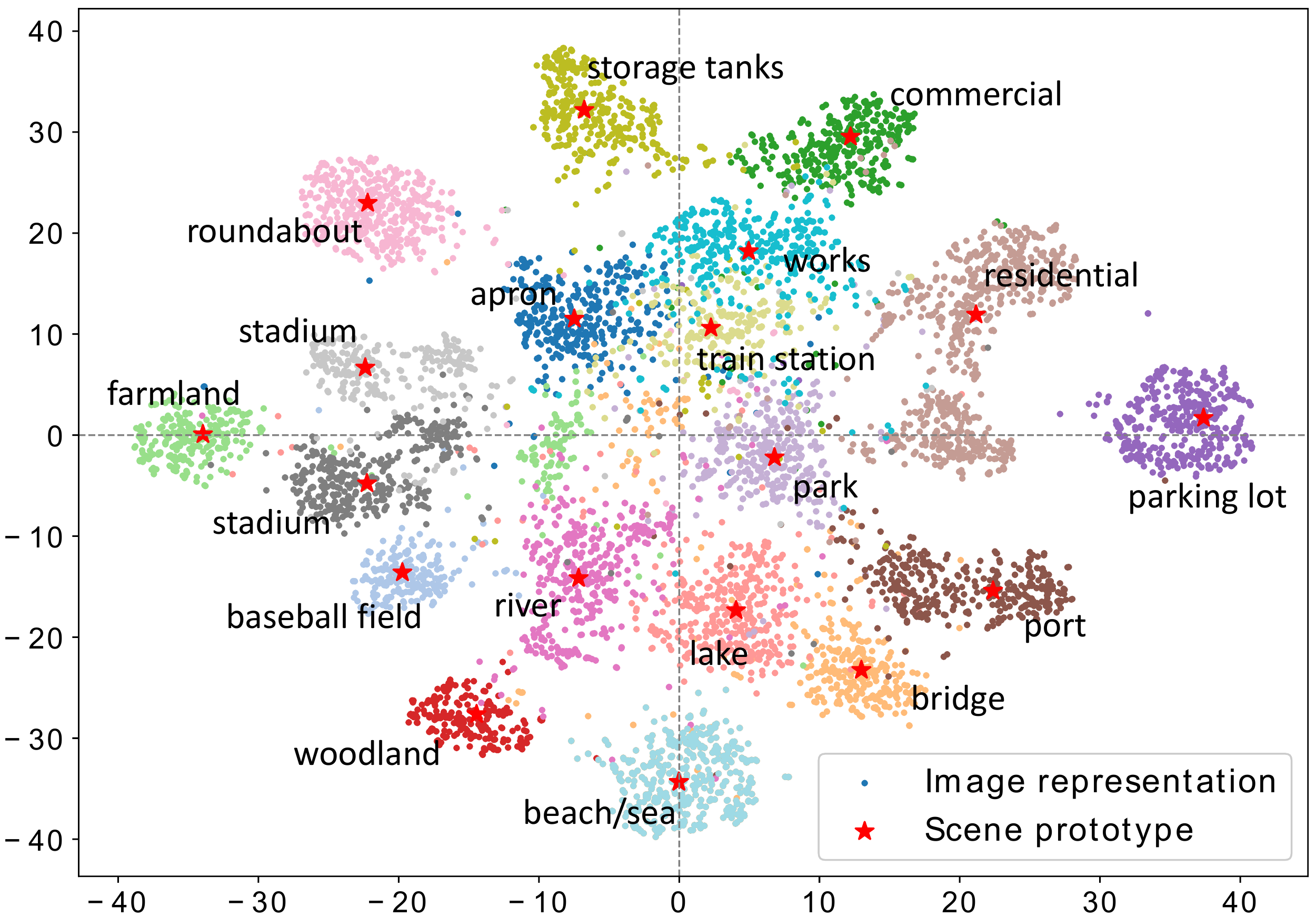}}
\caption{T-SNE visualization of image representations and scene prototypes learned by VGGNet on (a) UCM and (b) AID datasets, respectively. Dots in the same color represent features of images belonging to the same scene, and stars denote scene prototypes.}
\label{fig:vis_tsne}
\end{figure}

Let’s start with the conclusion: exploiting single-scene images significantly contributes to our task. To analyze its benefit, we mainly compare CNNs* and CNNs. It can be observed that even with identical network architectures, the performance of CNN is superior to that of CNN*. More specifically, VGGNet achieves the highest improvement of the mean $F_1$ scores, $19.26\%$, in comparison with VGGNet*. NASNet shows higher performance in all metrics compered to ResNet*, while other CNNs perform poorly in only the mean example-based precision with respect to their corresponding CNNs*. Besides, we visualize features of single-scene images learned by VGGNet on UCM and AID datasets via t-SNE, respectively. As shown in Figure~\ref{fig:vis_tsne}, extracted features are discriminative and separable in the embedding space, which demonstrates the effectiveness of learning the embedding function on single-scene aerial image datasets. To summarize, except for learning scene prototypes, single-scene training samples can also benefit multi-label scene interpretation by pretraining CNNs which are further utilized to initialize the embedding function. 

We exhibit several example predictions of PM-ResNet trained on UCM2MAI in Table \ref{tab:predictions_ucm2MAI}. False positives are marked as red, while false negatives are in blue. As shown in the forth example at the top row, we see that PM-Net can accurately perceive aerial scenes even in complex contexts, but unseen scene appearance (i.e. apron and runway in snow) can influence its prediction.

\subsection{Results on AID2MAI}
Table~\ref{tab:aid2MAI} reports numerical results on the AID2MAI configuration. It can be seen that the performance of PM-Net is superior to all competitors in the mean $F_1$ score. Compared to Mem-N2N-VGGNet, the proposed PM-VGGNet increases the mean $F_1$ and $F_2$ scores by $6.70\%$ and $7.56\%$, respectively, while improvements reach $6.07\%$ and $0.64\%$ in comparison with VGGNet. PM-ResNet achieves the best mean $F_1$ score and example-based precision, $57.42\%$ and $70.62$, respectively. With NASNet as the backbone, exploiting the proposed memory retrieval module contributes to increments of $1.03\%$ and $1.71\%$ in mean $F_1$ and $F_2$ scores compared to directly learning NASNet on a small number of multi-scene samples.

We present some example predictions of PM-ResNet in Table~\ref{tab:predictions_aid2MAI}. As shown in the top row, PM-ResNet learned with a limited number of annotated multi-scene images can accurately identify various aerial scenes even image contextual information is complicated. The bottom row shows some inaccurate predictions. It can be observed that although bridge and parking lot account for relatively small areas in last two examples at the top row, the proposed PM-Net can successfully detect them. Similar observations can also be found in the first and third example at the bottom row that residential and parking lot are recognized by our network, even they are located at the corner. In conclusion, quantitative results illustrate the effectiveness of our network in learning to perform unconstrained multi-scene classification, and example predictions further demonstrate it.

\section{Conclusion}
\label{sec:conclusion}
In this paper, we propose a novel multi-scene recognition network, namely PM-Net, to tackle both the problem of {aerial scene classification in the wild} and scarce training samples. To be more specific, our network consists of three key elements: 1) a prototype learning module for encoding prototype representations of variant aerial scenes, 2) a prototype-inhabiting external memory for storing high-level scene prototypes, and 3) a multi-head attention-based memory retrieval module for retrieving associated scene prototypes from the external memory for recognizing multiple scenes in a query aerial image. For the purpose of facilitating the progress as well as evaluating our method, we propose a new dataset, MAI dataset, and experiment with two dataset configurations, UCM2MAI and AID2MAI, based on two single-scene aerial image datasets, UCM and AID. In scene prototype learning, we train the embedding function on most of single-scene images as we aim to simulate the real-life scenario, where massive single-scene samples can be collected at low cost by resorting to OSM data. To learn memory retrieval, our network is fine-tuned on only around 100 training samples from the MAI dataset. Experimental results on both UCM2MAI and AID2MAI illustrate that learning and memorizing scene prototypes with our PM-Net can significantly improve the classification accuracy. The best performance is achieved by employing ResNet as the embedding function, and the best mean $F_1$ score reaches nearly 0.6. We hope that our work can open a new door for further researches in a more complicated and challenging task, multi-scene interpretation in single images. Looking into the future, we intend to apply the proposed network to the recovery of weakly supervised scenes.

\section*{Acknowledgements}
This work is jointly supported by the European Research Council (ERC) under the European Union's Horizon 2020 research and innovation programme (grant agreement No. [ERC-2016-StG-714087], Acronym: \textit{So2Sat}), by the Helmholtz Association
through the Framework of Helmholtz AI [grant  number:  ZT-I-PF-5-01] - Local Unit ``Munich Unit @Aeronautics, Space and Transport (MASTr)'' and Helmholtz Excellent Professorship ``Data Science in Earth Observation - Big Data Fusion for Urban Research'' and by the German Federal Ministry of Education and Research (BMBF) in the framework of the international future AI lab "AI4EO -- Artificial Intelligence for Earth Observation: Reasoning, Uncertainties, Ethics and Beyond" (Grant number: 01DD20001)
\bibliographystyle{elsarticle-num}
\bibliography{reference}

\end{document}